\documentclass{egpubl}
\usepackage{eg2021}
 
 \ConferencePaper        %

\usepackage[T1]{fontenc}
\usepackage{dfadobe}  
\usepackage{soul}

\usepackage{cite}  %
\BibtexOrBiblatex
\electronicVersion
\PrintedOrElectronic
\ifpdf \usepackage[pdftex]{graphicx} \pdfcompresslevel=9
\else \usepackage[dvips]{graphicx} \fi

\usepackage{egweblnk} 

\usepackage[utf8]{inputenc} %
\usepackage[T1]{fontenc}    %
\usepackage{hyperref}       %
\usepackage{url}            %
\usepackage{booktabs}       %
\usepackage{amsfonts}       %
\usepackage{nicefrac}       %
\usepackage{microtype}      %

\usepackage{xspace}
\usepackage{enumerate}
\usepackage{graphicx}
\usepackage{sidecap}

\usepackage{amssymb}
\usepackage{wrapfig}

\usepackage{amssymb, bm}
\usepackage{amsmath}
\usepackage{siunitx}
\usepackage{microtype}

\DeclareMathOperator*{\argmax}{argmax} %
\DeclareMathOperator*{\argmin}{argmin} %

\newcommand\norm[1]{\left\lVert#1\right\rVert}

\newcommand{\ours}{NGP\xspace}
\newcommand{\oursbp}{NGP-BP\xspace}
\newcommand{\oursplus}{NGP-plus\xspace}

\newcommand{\Dnorm}{D_\text{n}}
\newcommand{\Ddiffa}{D_\text{da}}
\newcommand{\Ddiff}{D_\text{df}}
\newcommand{\Ddepth}{D_\text{depth}}
\newcommand{\Dnoc}{D_\text{noc}}

\newcommand{\Gshape}{\mathcal{G}_\text{shape}}
\newcommand{\zshape}{z_\text{s}}  %

\newcommand{\Gdepthcomp}{G_{d \to c}}
\newcommand{\Gcompdepth}{G_{c \to d}}

\newcommand{\Gprop}{\mathcal{G}_\text{prop}}

\newcommand{\Gdepth}{\mathcal{G}_\text{depth}}
\newcommand{\Gnorm}{\mathcal{G}_\text{n}}
\newcommand{\Gdiffa}{\mathcal{G}_\text{da}}
\newcommand{\Gspec}{\mathcal{G}_\text{sa}}
\newcommand{\Grespec}{\mathcal{G}_\text{sp}^\text{real}}
\newcommand{\Grough}{\mathcal{G}_\text{r}}
\newcommand{\Ediffa}{E_\text{da}}

\newcommand{\Rdiff}{\mathcal{R}_\text{diff}}
\newcommand{\Rbp}{\mathcal{R}_\text{BP}}

\newcommand{\NOC}{noc}

\newcommand{\zdiffa}{z_\text{da}}  %
\newcommand{\dnoc}{\bm{d}_\text{noc}}

\newcommand{\Lossdepth}{L_\text{depth}}
\newcommand{\Lossdiff}{L_\text{df}}
\newcommand{\Lossdiffa}{L_\text{da}}
\newcommand{\Lossnorm}{L_\text{n}}
\newcommand{\Lossnoc}{L_\text{noc}}
\newcommand{\Losszdiffa}{L_{z_\text{da}}}
\newcommand{\LossKL}{L_\text{KL}}
\newcommand{\LosstwoDmodeling}{L^{\text{2D}}_\text{modeling}}
\newcommand{\LossthreeDmodeling}{L^\text{3D}_\text{modeling}}

\newcommand{\GaussianDistribution}{\mathcal{N}(0, \mathcal{I})}

\newcommand{\todo}[1]{{\color{red}{[TODO: #1]}}}
\newcommand{\name}{Neural Graphics Pipeline\xspace}
\newcommand{\namesmall}{NGP\xspace}

\usepackage{color}
\definecolor{blue}{rgb}{0,0,1}
\definecolor{red}{rgb}{1,0,0}
\definecolor{green}{rgb}{0,.5,0}
\definecolor{orange}{rgb}{0.75, 0.4, 0}
\newcommand{\xl}[1]{{\color{magenta}\textbf{}#1}\normalfont}
\newcommand{\xlc}[1]{{\color{magenta}\textbf{xl:}#1}\normalfont}

\newcommand{\dc}[1]{{\color{red}\textbf{}#1}\normalfont}
 
\newcommand{\new}[1]{{\color{black}\textbf{}#1}\normalfont}
\newcommand{\rev}[1]{{\color{black}\textbf{}#1}\normalfont}

\title{
Towards a \name for\\ Controllable Image Generation 
}

\author[X. Chen \& D. Cohen-Or \& B. Chen \& N. Mitra]
{\parbox{\textwidth}{ \centering
        Xuelin Chen$^{1,2}$\hspace{.5cm}
        Daniel Cohen-Or$^{3}$\hspace{.5cm}
        Baoquan Chen$^{1,4}$\hspace{.5cm}
        Niloy J. Mitra$^{2,5}$\hspace{.5cm}
        }
        \\
{\parbox{\textwidth}{ \centering
         $^1$CFCS, Peking University\hspace{.5cm}
         $^2$University College London\hspace{.5cm}
         $^3$Tel Aviv University\hspace{.5cm}
         $^4$AICFVE, Beijing Film Academy\hspace{.5cm}
         $^5$Adobe Research\hspace{.5cm}
       } 
}
}

\begin{document}

\teaser{
    \centering
    \includegraphics[width=\textwidth]{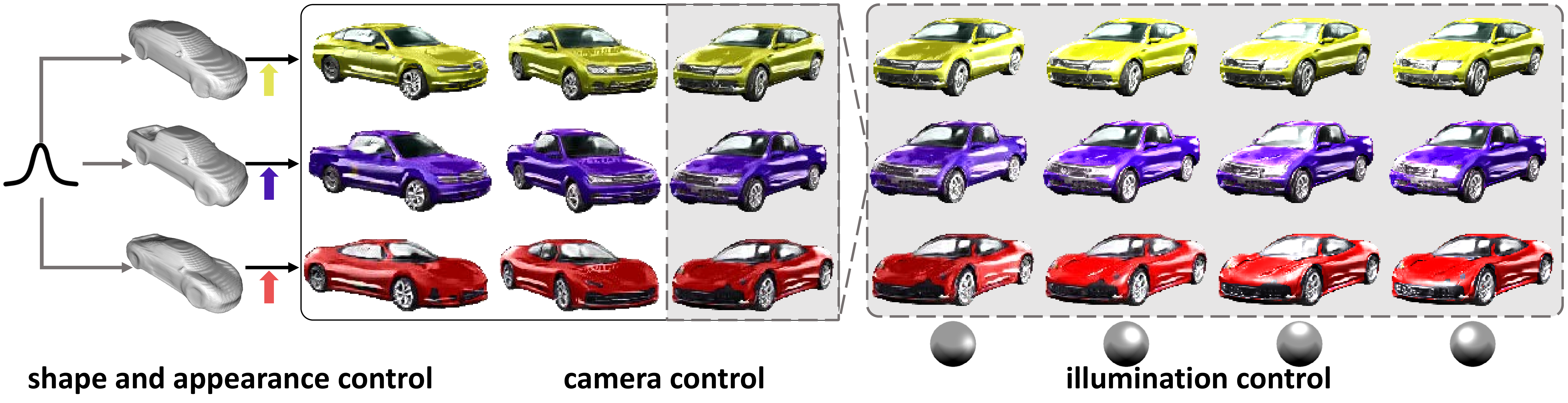}
    \caption{ 
      We present a \name, a GAN-based model that samples a coarse 3D model,  
      provides direct control over camera and illumination, and responds to geometry and appearance edits. 
      NGP is trained directly on unlabelled real images. 
      Mirrored balls~(right-bottom) indicate corresponding illumination setting.  
      }
\label{fig:teaser}
}

\maketitle

\begin{abstract}
    In this paper, we leverage advances in neural networks towards forming a neural rendering for controllable image generation, and thereby bypassing the need for detailed modeling in conventional graphics pipeline.
To this end, we present \textit{\name} (\namesmall), a hybrid generative model that brings together neural and traditional image formation models. 
\namesmall decomposes the image into a set of interpretable appearance feature maps, 
uncovering direct control handles for controllable image generation.
To form an image, NGP generates coarse 3D models that are fed into neural rendering modules to produce view-specific interpretable 2D maps, which are then composited into the final output image using a traditional image formation model.
Our approach offers  control over image generation by providing direct handles controlling illumination and camera parameters, in addition to control over shape and appearance variations.
The key challenge is to learn these controls through unsupervised training 
that links generated coarse 3D models with unpaired real images via neural and traditional (e.g., Blinn-Phong) rendering functions,  without establishing an explicit correspondence between them.
We demonstrate the effectiveness of our approach on controllable image generation of single-object scenes.
We evaluate our hybrid %
modeling framework, compare with neural-only generation methods (namely,  DCGAN, LSGAN, WGAN-GP, VON, and SRNs), report improvement in FID scores against real images, and demonstrate that   NGP supports direct controls common in traditional forward rendering.  
Code is available at \url{http://geometry.cs.ucl.ac.uk/projects/2021/ngp}.


\end{abstract}  

\section{Introduction}

Computer graphics produces images by \textit{forward rendering} 3D scenes. While this traditional approach provides \textit{controllability} in the form of direct manipulation of camera, illumination, and other rendering parameters, the main bottleneck of the classic approach is content creation, that is the explicit need to author detailed scenes.  
Neural networks have recently given raise to \textit{neural rendering} as an alternative approach wherein specialized networks are trained end-to-end to operate on deep features stored on sparse geometry (e.g., voxels~\cite{dcgan,deepvoxels,rendernet,hologan,neuralrerendering,von,neuralRepoNet_19}, points~\cite{aliev2019neural}, surface patches~\cite{atlasnet}) to directly produce pixel colors. 
Neural rendering revolutionizes  image synthesis workflow by  bypassing the content creation stage,  however, they lack the level of controllability supported in traditional rendering.

\begin{figure*}[t!]
    \centering
	\includegraphics[width=1.0\textwidth]{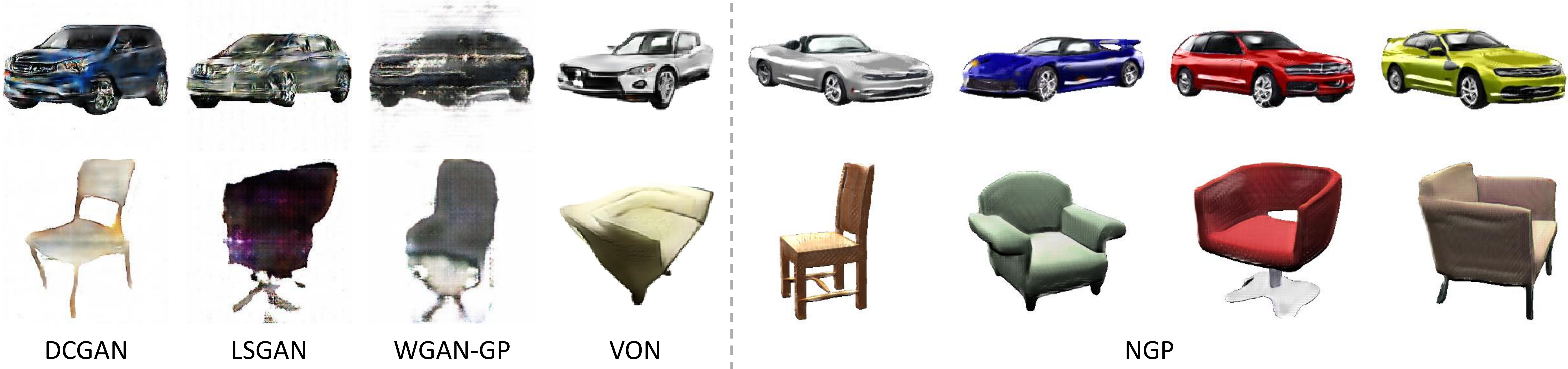}
    \caption{
    Left: image samples generated from existing approaches on cars and chairs categories; Right: image samples generated from NGP. The models have all been trained on the same datasets. 
    }
    \label{fig:early_comparison}
\end{figure*}

\begin{figure*}[t]
    \centering
	\includegraphics[width=1.0\textwidth]{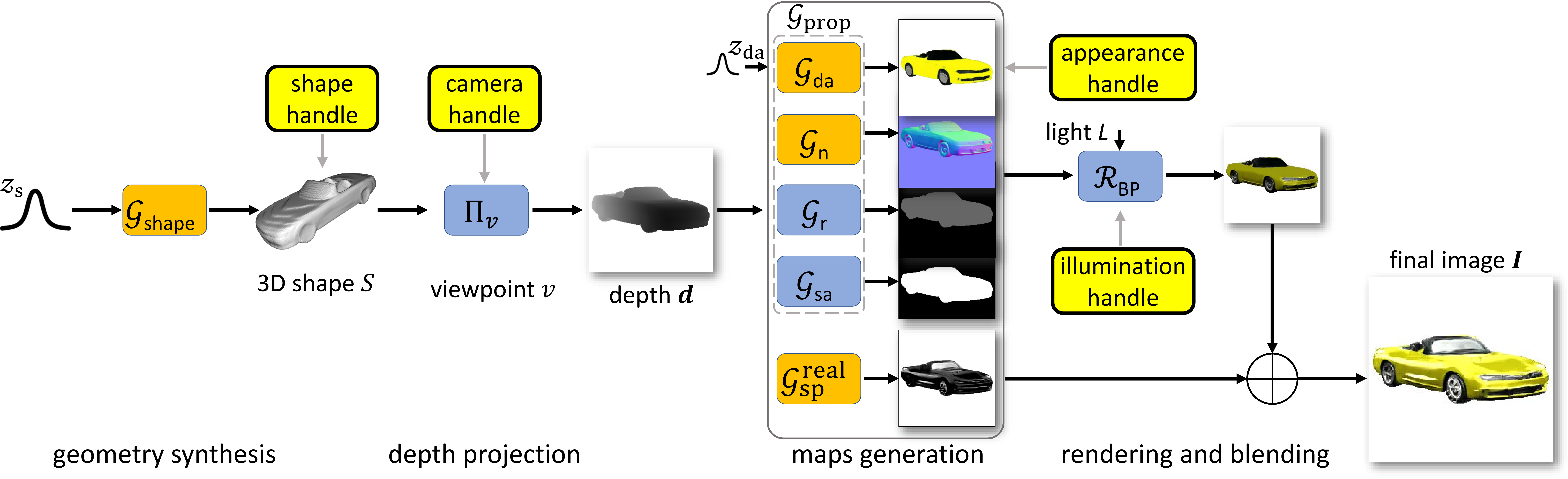}
    \caption{
    \textbf{NGP at inference time.} 
    At test time, starting from a sampled noise vector $\zshape$ and a set of user control signals (marked in yellow), NGP 
     uses a combination of learned networks (marked in mustard) 
    and fixed functions (marked in blue) to produce a range of interpretable feature maps, which are then combined to produce a final image $\bm{I}$. 
    }
    \label{fig:neural-modeling}
\end{figure*}

\new{
In this work, we leverage advances in neural networks towards forming a neural rendering for controllable image generation, and thereby bypassing the need for detailed modeling in conventional graphics pipeline.
As a first attempt along this avenue, }
we introduce \textit{\name}~(NGP), 
a hybrid generative approach that uses neural network to produce coarse 3D content, decorated with view-specific interpretable 2D features, that can then be consumed by traditional image formation models --- see Figure~\ref{fig:neural-modeling}.
The approach relaxes the need for modeling a fully detailed scene model, while retaining the same traditional direct control over illumination and camera, in addition to controls over shape and appearance variations \new{--- see the controllable image generation results via these control handles in Figure~\ref{fig:teaser}}.

 NGP (see Figure~\ref{fig:neural-modeling}) consists of four modules: 
 (i)~a GAN-based generation of a coarse 3D model, 
 (ii)~a projection module that renders the coarse geometry into a 2D depth map,  
 (iii)~a set of \new{neural networks} to produce image-space interpretable appearance features (i.e., normal, diffuse albedo, specular map, roughness), %
 and (iv) a 2D renderer that takes these appearance maps along with user-provided conventional illumination (i.e., light positions with intensity) to produce the final images.

Training NGP is challenging because there is no direct supervision available in terms of paired or unpaired input and corresponding 2D interpretable   features. 
We present an \textit{unsupervised} learning setup for the proposed neural modeling framework.
Note that by generating interpretable intermediate maps, we link the 3D and 2D images without any explicit correspondence information between them.
The core of NGP consists of a network that parameterically translates a depth image to an image with realistic appearance. 
These additional parameters, which disambiguate the translation, are in fact the handles that controls the image generation of the trained network. A notable feature of NGP, which is based on unsupervised unpaired training, is the ability of collectively learn from synthetic data and real images.

\new{By incorporating knowledge from graphics pipeline into neural image synthesis, 
we demonstrate the effectiveness of our approach for controllable image generation results of single-object scenes.}
We extensively evaluate our hybrid modeling framework against several competing neural-only image generation approaches~\cite{dcgan, lsgan, wgan, von, neuralRepoNet_19}, 
rate the different methods using the established FID score~\cite{heusel2017gans, lucic2018gans} \new{(see a preview in Figure~\ref{fig:early_comparison})},
and present ablation studies to show the importance of our design choices. 
Our tests demonstrate the superiority of our method (i.e., lower FID scores) compared to other state-of-the-art alternatives, on both synthetic and real data. %

\section{Related Work}

\textbf{GAN-based image generation.}
Since the introduction of Generative Adversarial Nets (GANs)~\cite{goodfellow2014generative}, many GAN variants~\cite{dcgan, lsgan, wgan, lrgan} have been proposed to synthesize images conditioned on control variables sample from a Gaussian distribution. State-of-the-art GAN-based methods are now able to generate images with high level of realism~\cite{bigGAN, wang2018high, stylegan}. 
While it is increasingly possible to provide guidance through conditional latent code~\cite{conditionalGAN, acgan}, structured latent space~\cite{blockgan, genesis, obj_comp, dtlcgan},  style example~\cite{gatys2016imagestyletransfer}, or semantic specifications~\cite{spade}, it still remains difficult to directly control generated imagery by updating all of geometry, camera, illumination, or material parameters. 
We were particularly inspired by the recently proposed visual object network~\cite{von} that takes a generated rough shape and trains a 2D texture network for adding texture to synthesize images. Different from ours, they directly output final RGB images, and do not provide access to 
interpretable intermediate features, and thus, prevents direct illumination control. %
We use unsupervised training, avoiding associating images with attributes or tags to allow scaling in terms of variety, richness, and realism.  

\textbf{3D generative neural networks.}
Researchers have also developed various generative networks for automatic content creation, ranging from single object generation~\cite{3dgan, im-net, deepsdf, grass, zhu_siga18, gao_siga19, mo2019structurenet, pointflow}, indoor scene synthesis~\cite{ma_siga18, wang2019planit, ritchie2019fast, yang2018automatic}, urban landscape and terrain  generation~\cite{KellyGuerreroEtAl_FrankenGAN_SigAsia2018,terrainGAN19}. 
The generated geometry, however, is still not sufficiently detailed and/or assigned with plausible materials, to be directly rendered by traditional forward rendering to produce high-quality images.

\textbf{Neural rendering.}
A particularly exciting breakthrough is neural rendering, where deep features are learned on coarse geometry (e.g., voxels, points), and then neurally rendered to produce a final image.
Most of the proposed approaches use supervised training and/or largely target novel view synthesis task~\cite{deepvoxels,thies2019deferred,neuralrerendering,rendernet, neuralavatar,aliev2019neural,zhang2018deep,neuralRepoNet_19, deepcg2real, tbn, mildenhall2020nerf}, with the output optionally conditioned using latent vectors (e.g., appearance vectors in \cite{neuralrerendering}). 
In the unsupervised setting, GQN~\cite{eslami2018neural_science} and HoloGAN~\cite{hologan} allow camera manipulation and model complex background clutter. However, since the learned features are deep, they cannot, yet, be manipulated using traditional CG controls. For example, one cannot freely control illumination in such an image generation pipeline.

\section{Formulation}
\label{sec:method}

\subsection{Overview}
Traditional computer graphics follows a \textit{model-and-render} pipeline, where a 3D scene is first modeled, and an image is then produced by rendering the 3D scene via a conventional renderer, a process that simulates the flow of light in physical world.
\new{While \namesmall follows a similar paradigm, it bypasses the need to directly model an elaborated scene with all essential assets in 3D for rendering.}
Figure~\ref{fig:neural-modeling} presents an overview of \namesmall at \textit{inference} time
: we first sample a coarse 3D shape using a neural network, followed by a set of learned generators producing view-specific interpretable reflectance property maps, along with a neural-rendered specular map. 
We assume the reflectance of the viewed content in the scene is characterized by a set of property maps: diffuse albedo, surface normal, monochrome roughness and specular albedo, which are then combined using the \textit{Blinn-Phong Reflection Model}.

\new{More specifically, to generate an image using \namesmall at inference time, }
a coarse shape $S$ is \new{first} generated from a latent shape code $S := \Gshape(\zshape)$, the shape is then projected from a viewpoint sample $v$ to form a 2D depth map $\bm{d} := \Pi_v(S)$.  The maps generation module then produces a set of intermediate maps, with $\zdiffa$ controlling the appearance (diffuse albedo).
The generated reflectance maps from $\Gprop := (\Gdiffa, \Gnorm, \Grough, \Gspec)$ are then fed into a fixed function \textit{Blinn-Phong} renderer $\Rbp$ (see Appendix~\ref{appendix:BP_function} in the appendix for details) to illuminate the viewed content under a given light setting $L$.
Blending the resultant rendering image with the realistic specular map generated by $\Grespec$, our image formation flow generates the final image  by sampling the space of ($\zshape$, $v$, $\zdiffa$, $L$) at inference time:
\begin{equation*}
    \small
        \bm{I}  := \Rbp \left(\Gprop(\Pi_v(\Gshape(\zshape)), \zdiffa), L \right) \oplus \Grespec(\Pi_v(\Gshape(\zshape))), 
    \label{eq:image_formation}
\end{equation*}
\new{where $\oplus$ denotes the image blending operation.}

\namesmall provides the user with several handles (highlighted in yellow in Figure~\ref{fig:neural-modeling}) to control the output image: 
(i)~a \textit{camera handle} offers direct control to rotate the camera view;  
(ii)~a \textit{illumination handle} offers direct control to specify the lights (position, intensity, and count); 
(iii)~a \textit{shape handle} to control the coarse geometry via direct editing and latent control; and 
(iv)~an \textit{appearance handle} to manipulate the appearance of the object via direct editing and latent control.  
The \namesmall is designed such that the output image meaningfully adapts to the user specifications. %

Next, we detail the individual modules in \namesmall and elaborate on how we train the networks without intermediate supervision. 
\new{As training data, 
we assume access to a collection of 3D shapes, a collection of reflectance maps, and a collection of real-world images for learning respective data prior.
Note that, similar to common GANs for image generation that usually assume no paired data (i.e., a random Gaussian sample is paired with a corresponding ground truth image for supervising the generator) available for training the networks, we do \textit{not} have the correspondences between the shape geometries, reflectance maps and final images, and the random samplings in the space of ($\zshape$, $v$, $\zdiffa$, $L$) for training our networks.
}

\begin{figure*}[t!]
    \centering
    \includegraphics[width=\textwidth]{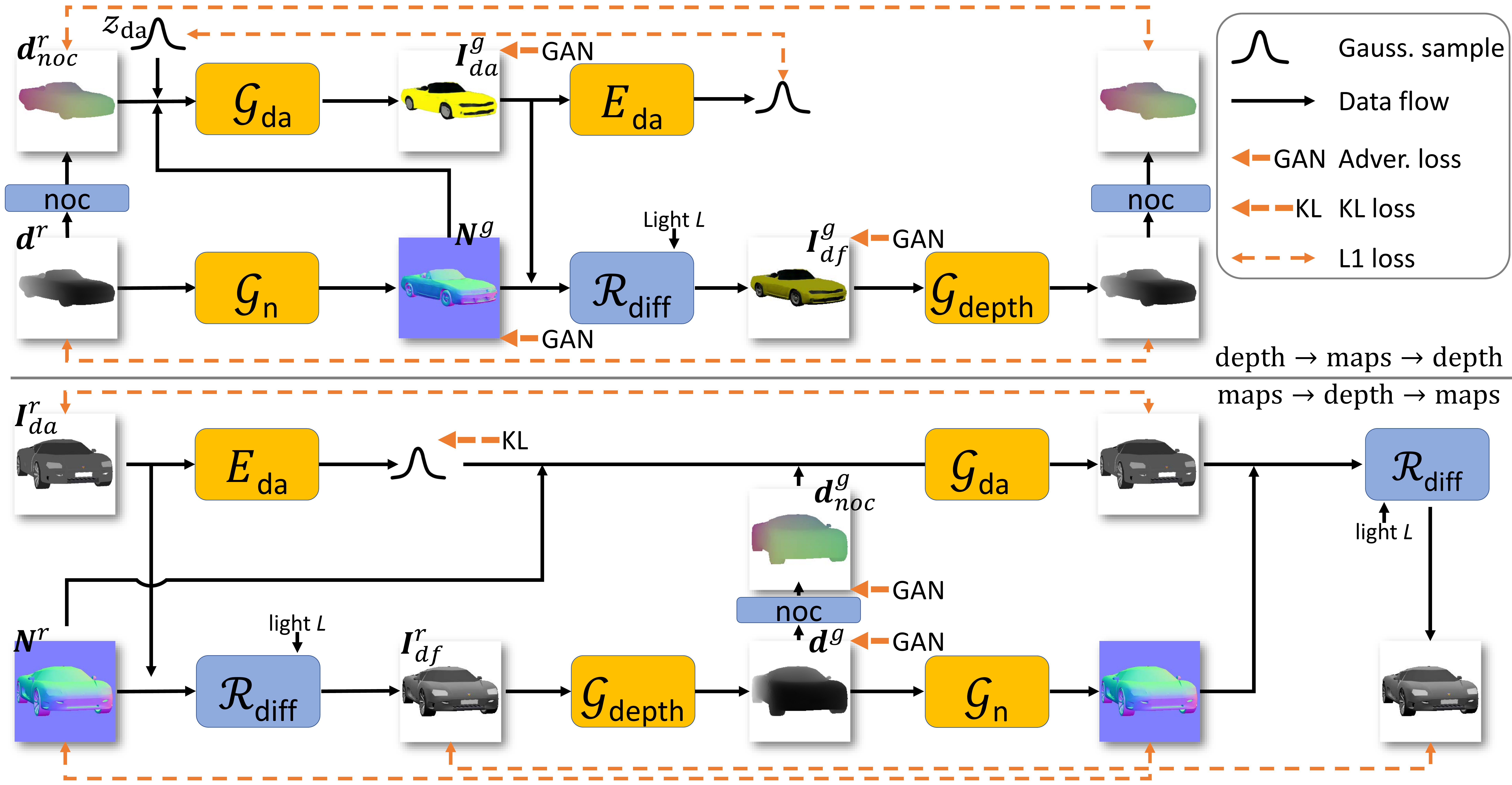}
    \caption{
    \textbf{Learning reflectance maps generation.} 
    The proposed architecture for training to jointly generate reflectance property maps from depth images using adversarial losses and cycle consistency losses to enable unpaired training. 
    Top: the cycle between the real depth image and the generated diffuse image composited from generated reflectance maps. 
    Bottom: the cycle between real diffuse image composited from real reflectance maps, and the generated depth maps. 
    }
    \label{fig:2D_network_architechture}
\end{figure*}

\subsection{Learning geometry synthesis}
We start with a category-specific 3D shape prior to capture rough geometry of the object, \textit{without} any reflectance properties. 
\new{A large collection of 3D shapes are used for training to learn this shape prior.}
We adopt the recently proposed IM-GAN~\cite{im-net}, which uses shape latent codes to produce implicit signed distance fields corresponding to realistic shapes, although alternate 3D generative models can  be used.

More specifically, we pretrain a 3D autoencoder with the implicit field decoder to produce a compact shape latent space %
for representing 3D shape implicit fields and use latent-GAN~\cite{achlioptas2017learning} on the trained shape latent space to produce realistic shape latent codes.
As a result, we learn a generator $\Gshape$ to map the Gaussian-sampled shape code $\zshape$ to a shape $S := \Gshape(\zshape)$. 

\subsection{Depth projection}
 Next, we project the coarse shape to 2D via a direct depth projection layer. Given a sampled shape $S$ and a sampled viewpoint $v$, which is parameterized by an extrinsic matrix $\bm{E} := \left[ \bm{R} | \bm{t} \right] \in \mathbb{R}^{3 \times 4}$ and camera intrinsics $\bm{K} \in {\mathbb{R}^{3 \times 3}}$, we obtain a coarse depth map $\bm{d}$ by projecting every \textit{visible} point $p$ (in homogeneous coordinates) on the surface $S$ as, 
$
    \bm{d} := \bm{K} \bm{E} p, \forall p \in S.
$
We use OpenGL calls for efficient rasterization and depth rendering. %
As we shall demonstrate, the original coarse depth map is itself sufficient for our end goal \new{--- image generation}. 
Although we train $\Gshape$ separately, if desired, \namesmall can be linked to a differentiable depth rendering layer and trained end-to-end.   

\subsection{Learning reflectance maps generation}
\label{subsec:map_generation}
Next, we elaborate on the modules to generate reflectance maps from a coarse depth map $\bm{d}$, including two constant function modules ($\Gspec$ and $\Grough$, for specular albedo and roughness maps, respectively) and two learned networks ($\Gdiffa$ and $\Gnorm$, for diffuse albedo and normal maps, respectively).

\textit{(i) Specular albedo and roughness maps generation.}
In absence of diverse specular albedo and roughness data to learn the data prior,
we simply realize $\Gspec$ and $\Grough$ as constant functions in the form:
$\Gspec(\bm{d}): \bm{I}^g_{sa} = c\mathcal{M}(\bm{d})$ and 
$\Grough(\bm{d}): \bm{\alpha}^g = \alpha\mathcal{M}(\bm{d}),$ 
where $\bm{I}^g_{sa}$ is the generated specular albedo map, $\bm{\alpha}^g$ the generated roughness map, $\mathcal{M}(\cdot)$ generates the mask of $\bm{d}$ by thresholding the depth, $c$ is a constant specular albedo (set to white) and $\alpha$ is a constant roughness (set to 4.0).

\textit{(ii) Learning to generate diffuse albedo and normal maps.} 
For \new{the training data to learn} $\Gdiffa$ and $\Gnorm$, we only have access to example `real' reflectance maps that comprise of real diffuse albedo maps $\mathbb{I}_{da}^r = \{ \bm{I}_{da}^r \}$ and detailed normal maps $\mathbb{N}^r = \{ \bm{N}^r \}$, 
along with corresponding viewpoints. %
Note that, given the light setting $L$ \new{(modelled as a set of white directional lights during training)}, each set of real reflectance maps, denoted by $(\bm{I}_{da}^r, \bm{N}^r)$, can be used to render a real diffuse image $\bm{I}^r_{df}$ using the diffuse reflection component (denoted as $\Rdiff$) in the Blinn-Phong equation.

Given the coarse depth image $\bm{d}$ and the viewpoint $v$ parameterized by $\bm{E}$ and $\bm{K}$, the task is then \new{training $\Gdiffa$ and $\Gnorm$ to} synthesize a pair of generated reflectance maps $(\bm{I}_{da}^g, \bm{N}^g)$ that can be used to render a diffuse image $\bm{I}^g_{df}$. 
Training with supervision would be relatively easy, and can be seen as a standard task. 
However, we do not have access to ground truth maps for supervision, i.e., the shape generated from the shape network comes \textit{without} any ground truth reflectance properties.
Hence, we treat this as an unpaired image-to-image translation problem. 
Our key idea is to do a cycle translation between the depth map and the diffuse image (i.e., th product of the diffuse albedo and detailed normal map), via the fixed diffuse rendering function $\Rdiff$. 
Specifically, we design a cycle-consistent adversarial network that \textit{jointly} generates $(\bm{I}_{da}^g,\bm{N}^g)$ from $\bm{d}$. 
Figure~\ref{fig:2D_network_architechture} shows the proposed architecture.

Given `real' depth maps $\mathbb{D}^r = \{\bm{d}^r\}$ produced from the depth projection, we train a network $\Gnorm$ to generate a detailed normal map $\bm{N}^g = \Gnorm(\bm{d}^r)$ that fools a discriminator trained to differentiate the real and generated detailed normal maps,
and another network $\Gdiffa$ to generate a diffuse albedo map $\bm{I}_{da}^g$ that fools a diffuse albedo map discriminator (Figure~\ref{fig:2D_network_architechture}-top). 
Note that we do \textit{not} enforce one-to-one mapping from depth maps to diffuse albedo maps, but rather condition the generation using random Gaussian sample code $\zdiffa$. 
In practice, we found the network $\Gdiffa$ difficult to train in absence of 3D object-space coordinates, as opposed to the view-dependent camera-space coordinates provided by the depth map. 
Hence, we use the intrinsic $\bm{K}$ and extrinsic $\bm{E}$ camera parameters, to enrich $\bm{d}^r$ to the normalized object coordinates~(NOC)~\cite{noc} system to obtain $\dnoc^r := \NOC(\bm{d}^r,\bm{K},\bm{E})$. 
Further, we found that the generated normal map $\bm{N}^g$ helps improving the generation of the diffuse albedo, as the detailed normal map provides more detailed geometry information. Therefore, we give  $\Gdiffa$ as input $\dnoc^r$, $\bm{N}^g$, and $\zdiffa$ resulting in: 
$
    \bm{I}_{da}^g := \Gdiffa(\dnoc^r, \bm{N}^g, \zdiffa)
    \label{eq:diffuse_from_G_diffuse}
$.
Following these two generation networks, a differentiable diffuse renderer $\Rdiff$ takes as input $\bm{N}^g$ and $\bm{I}_d^g$ to generate a diffuse image $\bm{I}^g_{df} := \Rdiff(\bm{N}^g, \bm{I}_{da}^g, L)$.

On the other end (Figure~\ref{fig:2D_network_architechture}-bottom), given the `real' diffuse albedo map $\bm{I}_{da}^r$ and detailed normal map $\bm{N}^r$, we introduce an encoder $\Ediffa$ to estimate a Gaussian-distributed diffuse albedo code from the real diffuse albedo map $\bm{I}^r_d$. 
In addition, a `real' diffuse image is rendered via $\bm{I}^r_{df} := \Rdiff(\bm{N}^r, \bm{I}_{da}^r, L)$, taken as input to the depth network $\Gdepth$ to generate a coarse depth map $\bm{d}^g = \Gdepth(\bm{I}^r_{df})$ that fools a coarse depth map discriminator.

We jointly train all the networks $\Gnorm$, $\Gdiffa$, $\Ediffa$, $\Gdepth$ with a set of adversarial losses and cycle-consistency losses, as illustrated with  the dashed arrows in Figure~\ref{fig:2D_network_architechture}.  
We also simultaneously train corresponding discriminators to classify the real from the generated maps/images. More details about the training losses can be found in the appendix (Appendix~\ref{appendix:losses}).
We use fixed light setting $L$ during training, placing \new{uniformly 4 overhead white directional} lights to light up the scene. 
Note that the light setting $L$ can be dynamically changed at inference time, resulting in illumination control in the generated images. %

\begin{figure}[t!]
    \centering
	\includegraphics[width=0.48\textwidth]{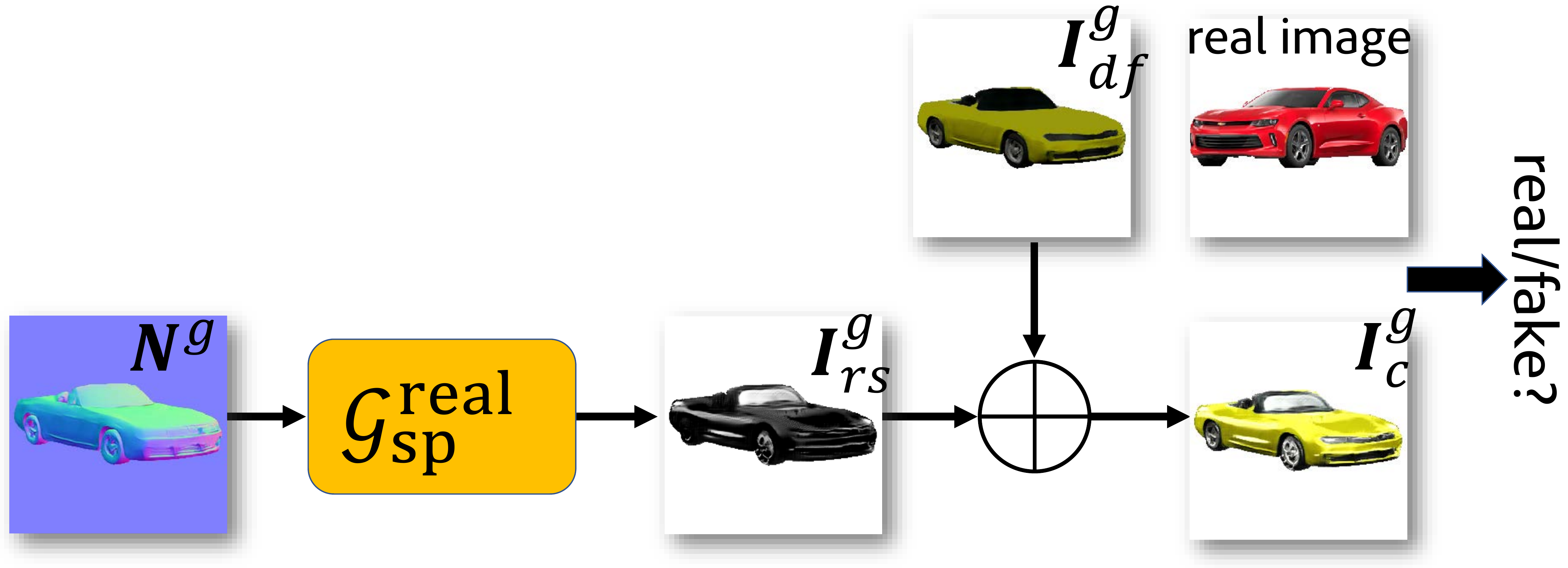}
    \caption{The cycle-consistent adversarial network for learning to generate realistic specular maps from normal maps using adversarial and cycle consistency losses. For simplicity, other essential modules in the training cycles are omitted.
    }
    \label{fig:real_network}
\end{figure}

\subsection{Learning realistic specular generation}
\label{subsec:real_spec}
To add further realism to the final image, we learn a realistic specular network $\Grespec$, which takes as input the generated detailed normal map $\bm{N}^g$, derived from the input depth, to generate a realistic specular map $\bm{I}^g_{rs}$. 
Blending this generated realistic specular map with the generated diffuse image $\bm{I}^g_{df}$ leads to a composite image $\bm{I}^g_{c}$ that fools a realistic images discriminator (see Figure~\ref{fig:real_network}) \new{, which is trained to differentiate the generated final composite images from the real-world images}.
\new{The training data here for learning $\Grespec$ is only a collection of real-world images.}
To enable training without paired data, we \new{again} designed a cycle-consistent adversarial network for learning $\Grespec$. 
\new{The key idea shares the same adversarial cycle-consistency idea as adopted in Section~\ref{subsec:map_generation}, and thus we do not expand on more details of the network architecture.}
Note the realistic specular generator can be linked to the networks of training reflectance map generators, making the setup end-to-end trainable. 
Note that this realistic specular generation, while offers furthur realism to the final image, is only conditioned on the view-specific input $\bm{N}^g$ and $\bm{I}_{df}^g$, 
\rev{thus it remains unaffected by the illumination specifications (i.e., uncontrollable) and can be inconsistent across views.
With the balance between realistic specular generation and loss of full control, we offer different ways of generating images with trade-off between the realism and control (Sec.~\ref{sec:exp}).
Further improving controllability, particular multi-view consistency of the realistic specular generation, is left to future research.}

\begin{figure*}[t]
    \centering
    \includegraphics[width=1.0\textwidth]{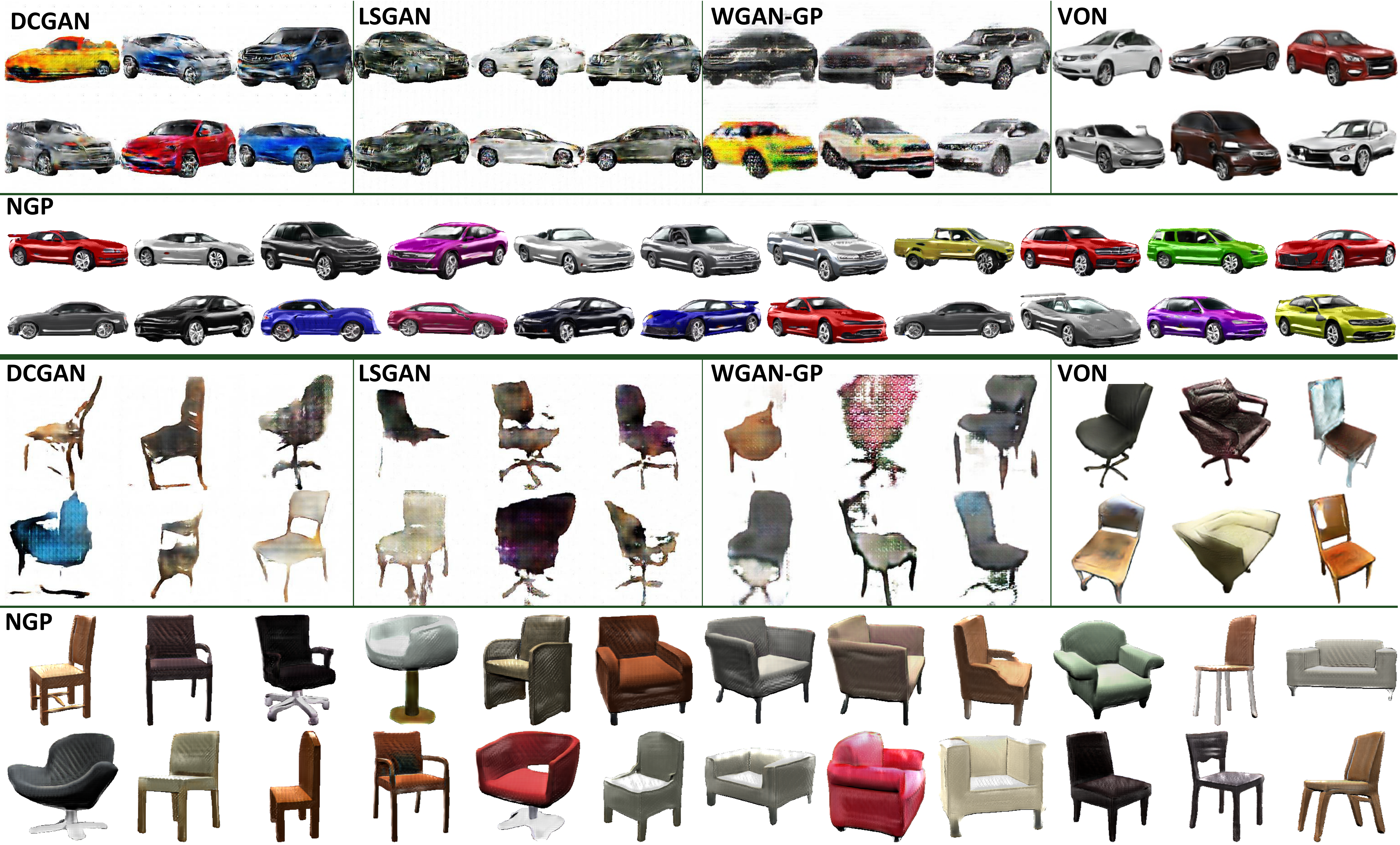}
    \caption{
    \textbf{Qualitative comparison with baselines.} 
    \ours versus DCGAN~\cite{dcgan}, LSGAN~\cite{lsgan}, WGAN-GP~\cite{wgan}, and VON~\cite{von}. All the models were trained on the same set of real-world images. %
    }
    \label{fig:visual_comparison}
\end{figure*}

\begin{figure*}[t]
    \centering
    \includegraphics[width=1.0\textwidth]{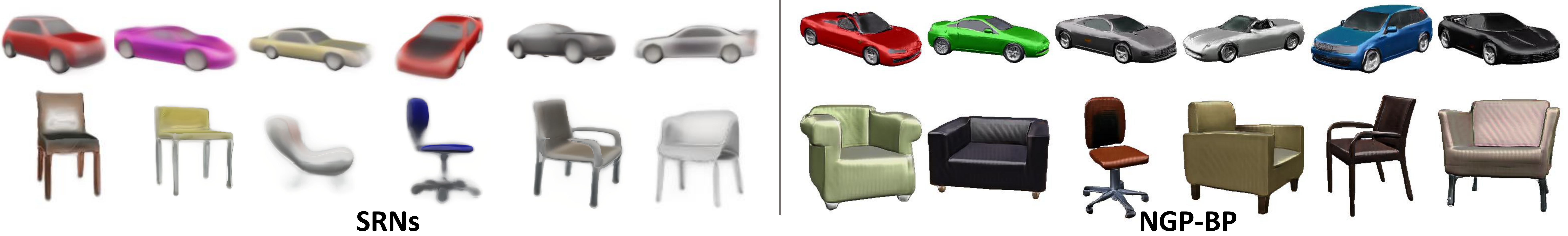}
    \caption{\textbf{Comparison with SRNs.} For fair comparison, we 
    give SRNs~\cite{neuralRepoNet_19} full camera information and use the depleted \oursbp option. Please refer to the appendix (Appendix~\ref{appendix:baseline_details}) for details. }
    \label{fig:srns_comprison}
\end{figure*}

\subsection{Implementation details}
\label{sec:imp_details}
Hyperparameters and full network architectures for NGP are detailed in the following. Training of the presented models took around 5 days per class on a single Nvidia GeForce GTX 1080. A single forward pass takes around 180 ms and 1 GB of GPU memory. Note that while training on real images, we found accurately modeling  perspective effects, instead of an orthogonal camera assumption, to be important.  

\textbf{\textit{3D shape network.}}
For the coarse shape synthesis network, we adopt the IM-GAN architecture from~\cite{im-net}. Both generator and the discriminator are constructed by two hidden fully-connected layers, and the Wasserstein GAN loss with gradient penalty is adopted to train the latent-GAN. We refer readers to the original paper of IM-NET for more in-depth technical details.

\textbf{\textit{2D detailing networks.}}
We use a perspective camera with a focal length of 50mm (35 film equivalent).
The 2D networks takes as input depth images of $256 \times 256$ resolution, which is also the size for all reflectance maps.
For 2D maps generation networks, we use the ResNet encoder-decoder~\cite{cyclegan,huang2018multimodal} for all map generators. In addition, we concatenate the diffuse code $\zdiffa$ to all intermediate layers in the encoder of $\Gdiffa$~\cite{von}, the generated detailed normal map $N^g$ is fused to the first layer of the encoder of $\Gdiffa$ by concatenation. The ResNet encoder~\cite{resnet} is used for constructing $\Ediffa$.
We use mid $(70 \times 70)$ and large $(140 \times 140)$ receptive field size (RFS) for all discriminators (except the diffuse albedo discriminator), as the generation of these maps relies on the mid-level and global-structure features extracted by the corresponding discriminators; we use small $(34 \times 34)$ RFS for the diffuse albedo discriminator, as the generation of the diffuse albedo needs only low-level features extracted by the corresponding discriminator, such as \textit{local} smoothness, purity, repetitive pattern and etc.\ of the albedo color, paying less attention to the global structure of the generated diffuse albedo. Finally, we use the least square objective as in LS-GAN~\cite{lsgan} for stabilizing the training.

\section{Experiments}
\label{sec:exp}

We introduce the datasets, evaluation metrics, and compare \name against competing GAN-based and/or neural rendering baselines. 
Further details can be found in the appendix (Sec.~\ref{appendix:datasets} and Sec.~\ref{appendix:baseline_details}).
We evaluate our generated images, both qualitatively and quantitatively, on publicly-available datasets. %
For comparisons, at test time, we use two variants of our method: 
(i)~{\em \ours}: as the default option, the final image is generated by blending the diffuse rendering of $\Rdiff$ \rev{(Blinn-Phong specular excluded)} under 4 base overhead lights (same setting as in training) with the realistic specular map; and 
(ii)~{\em \oursbp}: as a depleted option, where we use the full Blinn-Phong renderer $\Rbp$ under 4 base lights overhead (same setting as in training), along with randomly sampled lights, but \textit{without} blending with the realistic specular map.

\begin{table}[b!]
    \centering
    \tiny
    \caption{FID comparison (lower is better) on real images data. Note that FIDs are computed against real images data.}
      
      \begin{tabular}{r|ccccccc}
        \toprule
            & DCGAN & LSGAN & WGAN-GP & VON  & \oursbp  & \ours \\
          \midrule
          car   &130.5 & 171.4 & 123.4 & 83.3 & \textbf{67.2} & \textbf{58.3} \\
          \midrule
          chair & 225.0 & 225.3 & 184.9 & 51.8 & \textbf{47.9} & \textbf{51.2}  \\
         \bottomrule
        \end{tabular}%
    \label{tab:fid_table_onreal}
\end{table}

\begin{figure*}[t!]
    \centering
    \includegraphics[width=1.0\textwidth]{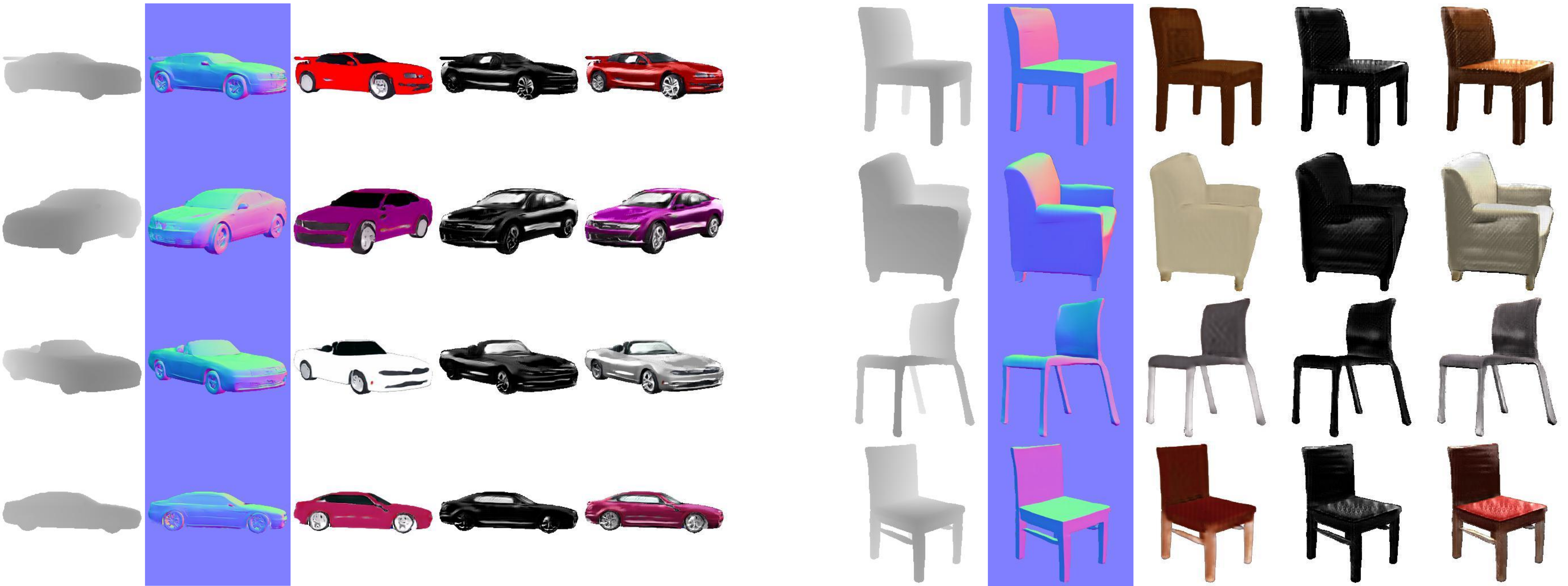}
    \caption{More results with generated intermediate maps. From left to right: input coarse depth map, generated detailed normal map, generated detailed diffuse albedo map, generated realistic specular map, and the final generated image of our method. The generated specular albedo map and roughness map by constant function generators are not shown here. 
    } 
    \label{fig:more_results_reflectance_maps} 
\end{figure*}

\subsection{Evaluations}
\label{sec:datasets}
\label{data_preparation}

\textbf{\textit{Datasets.}}
Our 3D dataset consists of chairs and cars from ShapeNet~\cite{chang2015shapenet} \new{for training 3D rough geometry generator}; 
as 2D datasets, we render each ShapeNet model in Blender~\cite{blender} to collect example real reflectance maps for training the reflectance map generators, while we use the real-world images dataset from VON~\cite{von}, which contains 2605 car images and 1963 chair images, for training the realistic specular generator.

\textbf{\textit{Baselines.}}
We compare our method against the following baseline methods:
DCGAN~\cite{dcgan}, LSGAN~\cite{lsgan}, WGAN-GP~\cite{wgan}, VON~\cite{von}, and SRNs~\cite{neuralRepoNet_19}, of which the details can be found in the appendix. 
Since SRNs assumes training images with full camera parameters, we train SRNs on Blinn-Phong rendered images with varying lighting. 
For a fair comparison, we compare separately to SRNs with \oursbp, reporting the FID computed against Blinn-Phong rendered images.

\textbf{\textit{Metrics.}}
Fr\'echet Inception Distance (FID) is an established measure comparing inception similarity score between distributions of generated and real images~\cite{heusel2017gans, lucic2018gans}.
To evaluate an image generation model, we calculate FID between the generated images set and a target real images set. Specifically, each set of images are fed to the Inception network~\cite{szegedy2015going} trained on ImageNet~\cite{deng2009imagenet}, then the features with length 2048 from the layer before the last fully-connected layer are used to calculate the FID. 
Lower FID score indicates image generation with better quality.

\textbf{\textit{Results.}}
We first compare our method against baseline methods (excluding SRNs) on the real-world images data.
Our method variants consistently outperform these baselines qualitatively and quantitatively.
In Table~\ref{tab:fid_table_onreal}, both \ours and \oursbp have the two best FID scores, outperforming other baseline methods by large margins. 
Qualitative comparisons on real images are presented in Figure~\ref{fig:visual_comparison}. 
Note the realism of specular highlights, the wheels and windscreens of the cars, or the varying illumination on the chairs. 
The GAN variants (i.e., DCGAN, LSGAN, and WGAN-GP) suffer from lower visual quality as they seek to directly map the Gaussian samples to final images, only producing results with roughly plausible content. 
\begin{wraptable}{r}{0cm}
\label{tab:fid_table_srns}
    \tiny
      \begin{tabular}{r|cc}
        \toprule
              & SRNs  & \oursbp \\
        \midrule
        car   & 167.0 & \textbf{30.0} \\
        \midrule
        chair & 50.3  & \textbf{32.0} \\
        \bottomrule
      \end{tabular}%
\end{wraptable} 
Among these variants, VON produces the closest results compared to \ours. 
Note that although our method provides control over illumination and camera, we do not see any performance degradation, but on the contrary, our method still produces slightly better visual results over VON.  
Interestingly,  observe that by imposing inductive bias on the image formation model used in traditional rendering, \oursbp results in superior quality results even when trained on the same dataset. 
We also conduct qualitative comparison with \oursbp against SRNs, as shown in Figure~\ref{fig:srns_comprison}, 
with quantitative numbers; See the inset table, note that FIDs are computed against Blinn-Phong images data.

\textbf{\textit{More results with intermediate maps.}}
In Figure~\ref{fig:more_results_reflectance_maps}, we present more qualitative results of generated images using \namesmall, along with the intermediate maps used to composite the final images.

\textbf{\textit{Ablation study.}}
Table~\ref{tab:fid_table_onreal} shows the ablation result of our realistic specular blending network $\mathcal{G}_\text{sp}^\text{real}$ using \oursbp \textit{v.s.} \ours.

\begin{figure*}[t!]
    \centering
    \includegraphics[width=\textwidth]{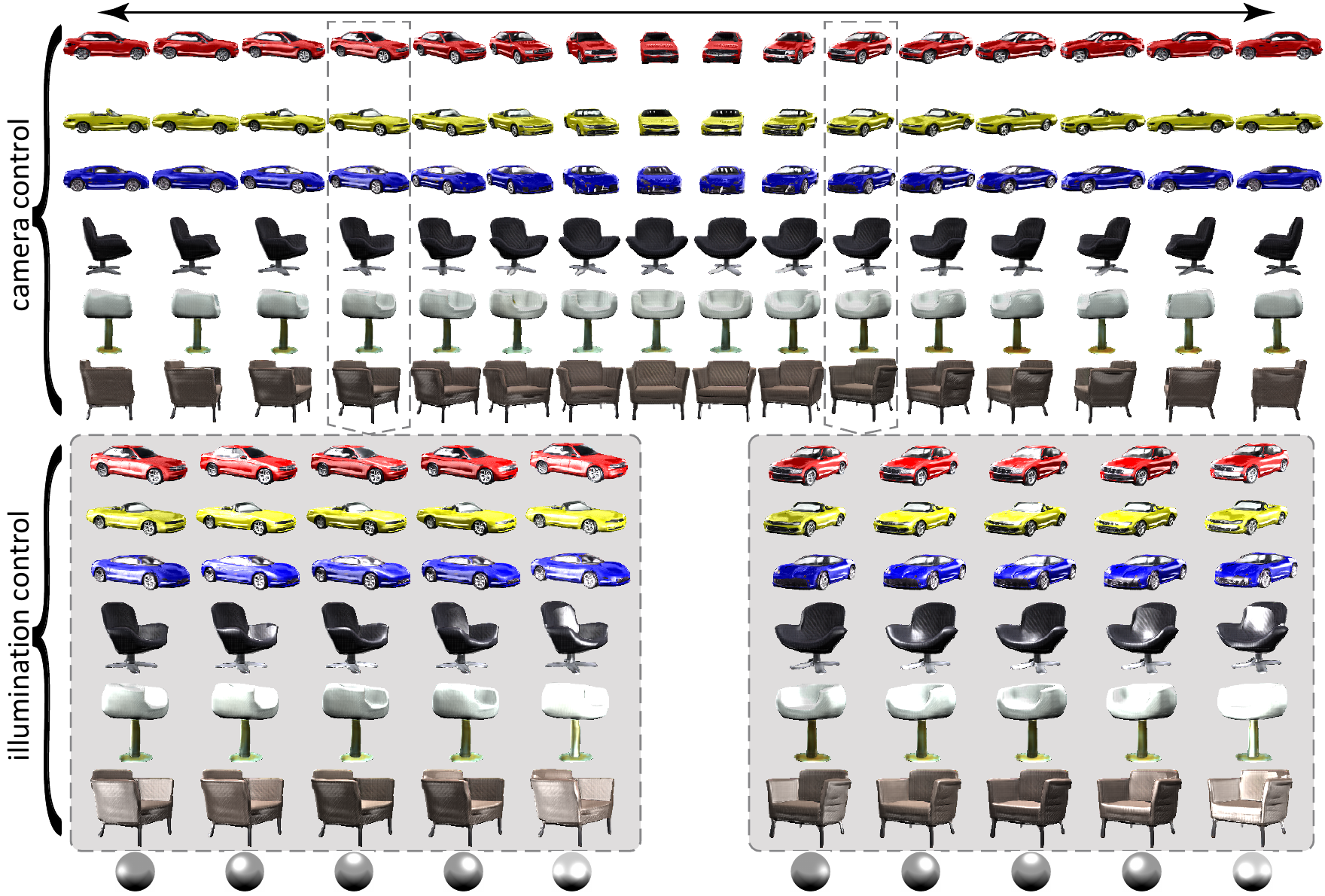}
    \caption{
    \textbf{Camera and illumination control.}
    (Top)~Camera control with object shape and appearance held fixed along rows. 
    For selected camera views~(marked at the top), we show (at the bottom) the corresponding generations under changing illumination (intensity, location, number of lights) as shown on the mirrored ball. Note the granularity of camera and illumination control enabled by ours. 
    }
    \label{fig:camera_illumination_control}
\end{figure*}

\begin{figure*}[]
    \centering
	\includegraphics[width=1.0\textwidth]{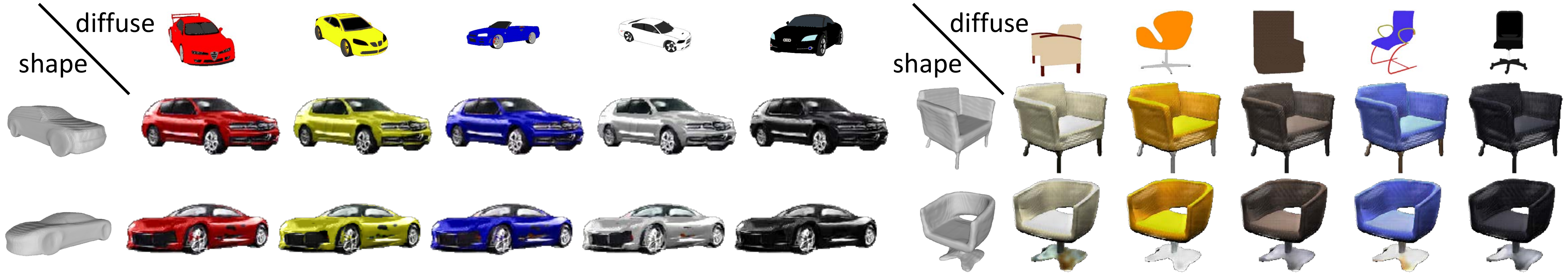}
    \caption{Appearance control via exemplar diffuse albedo images (top rows). Note that the specular highlights on the surface are preserved even under changes to the color of the cars/chairs.
    }
    \label{fig:appearance_control}
\end{figure*}

In addition, based on the default option NGP, we also conduct ablation studies to show the importance of the detailed normal map generator, the diffuse albedo map generator and the realistic specular generator in generating the final images. The quantitative results are presented in Table~\ref{table:ablation_study}, where:
\begin{enumerate}[(i)]
    \item NGP-w/o-$\Gnorm$ disables the detailed normal map generator in \ours, and uses the coarse normal map derived from the input coarse shape for the final image generation.
    \item NGP-w/o-$\Gdiffa$ disables the diffuse albedo generator in \ours, and uses a white diffuse albedo map for the final image generation.
    \item NGP-w/o-$\Grespec$ disables the realistic specular generator in \ours, such that the final image is produced without blending with the realistic specular map.
\end{enumerate}
\begin{table}[h!]
    \centering
    \small
    \caption{Ablation study shows the importance of each generator in generating the final images. \new{FIDs (lower is better) computed against real images data are reported in the table.}}
      
      \begin{tabular}{r|cccc}
        \toprule
            & \ours & NGP-w/o-$\Gnorm$ & NGP-w/o-$\Gdiffa$ & NGP-w/o-$\Grespec$  \\
          \midrule
          car   & \textbf{58.3} & 64.6 & 114.1 & 74.8 \\
          \midrule
          chair & \textbf{51.2} & 55.7 & 71.3 & 52.9 \\
         \bottomrule
        \end{tabular}%
\label{table:ablation_study}
\end{table}

\subsection{Controllable Image Generation}
\label{sec:applications}
The key advantage of NGP is retaining the controls available in traditional   modeling-and-rendering based image generation.
In the following, we demonstrate the various controls supported by our method. 
See Figure~\ref{fig:neural-modeling} and supplemental video. 

\textbf{\textit{Shape control.}}
NGP generates images of diverse shapes with ease via simply changing the shape code $z_\text{s}$. Additionally, the user can directly edit the coarse geometry, as shown in the video. 

\textbf{\textit{Camera control.}}
We also support \new{direct} camera view control for the final generated image while keeping all other factors.  Figure~\ref{fig:camera_illumination_control} illustrates the effect of changing the camera view for generating different final images. Note that earlier works including VON~\cite{von}, SRNs~\cite{neuralRepoNet_19}, and HoloGAN~\cite{hologan} also support various levels of camera control. 

\textbf{\textit{Illumination control.}}
Our method models detailed normal maps in the reflectance property maps generation stage, 
so that additional lights can be added on top with explicit control of the illumination (see  Figures~\ref{fig:neural-modeling}, \ref{fig:camera_illumination_control}). 
We call this more versatile option \oursplus (see details in Appendix~\ref{appendix:baseline_details}). 
Such level of control (i.e., explicit light count, position, and intensity) is not supported by VON~\cite{von} and HoloGAN~\cite{hologan}.
Figure~\ref{fig:camera_illumination_control} shows the effect of generating various images with different additional light settings.

\textbf{\textit{Appearance control.}}
The overall appearance, particularly the colorization, of the content in the generated images can be easily changed by providing an exemplar image as guidance, leading to controllable and various appearance in generated images (see Figure~\ref{fig:appearance_control}). 
Further, this allows the user to simply edit the diffuse albedo, akin to traditional control, using existing imaging tools, and render the final image using NGP, thus benefiting from the appearance disentanglement.

\subsection{Limitations}
\new{
While we present a solution towards incorporating knowledge from graphics pipeline into neural rendering, 
the quality of the generated images is nevertheless below the current limits of traditional computer graphics pipelines (e.g., using Blender) 
due to the insufficient image resolution and the lack of global illumination, shadows, and the ability to composite multiple objects for more complex scenes.
}

\rev{Moreover}, as has been mentioned in Section~\ref{subsec:real_spec}, the realistic specular generation lacks of control in the proposed preliminary solution since it is only conditioned on the view-specific normal map input and thus remains unaffected by illumination specifications and has  multi-view inconsistency.
\rev{
We also observed multi-view inconsistency from the appearance of objects in the results.
This is caused by modeling/texturing in 2D instead of 3D, which is in contrast to recent works that directly model the scene in 3D via implicit neural fields, and further leads to the inability to generalize to out-of-distribution cameras.
Also note that we use the inductive bias of the traditional rendering model (i.e., Blinn-Phong model), which, while allows more control handles, leads to somewhat unrealistic specular highlights characteristic of the Blinn-Phong model. 
This is a trade-off --- NGP(-plus) offers realistic specular highlights while losing absolute control over it, while NGP-BP has the best control over the specular highlights, while being somewhat unrealistic.
}

While we acknowledge these shortcomings, we believe this will rapidly change, as we have witnessed in the context of GANs in general or neural rendering in particular. 
One axis of improvement, will be targeting larger image sizes (e.g., $1024 \times 1024$ instead of current $256 \times 256$), possibly using a progressive GAN setup. 
\new{Another orthogonal axis of improvement, which we believe is of higher priority, will be boosting the rendering quality by incorporating more advanced but fully differentiable graphics rendering models instead of the Blinn-Phong model. 
This will also help to restore the control for the realistic specular generation.}
Last, our models are class-specific \new{and limited to single-object scenes}, thus, currently separate networks need to be trained for new shape categories \new{and can not synthesize images with multiple objects}. However, conceptually, being unsupervised, we believe that \name can be used across many classes, as long as we can get sufficient data volumes for training, and stronger networks with larger capacities \new{for compositing multiple objects}.

\section{Conclusions and Future Work}

We have presented a novel graphics pipeline that combines the strength of neural networks with the effective direct control offered in traditional graphics pipeline. 
We enable this by designing neural blocks that generate coarse 3D geometry and produce interpretable 2D feature layers,  that can then be directly handled by a fixed rendering function to produce a final image. It is important to emphasize that our training is completely unsupervised and no attributes like texture or reflectance indices are associated with the images. This allows using both real and synthetic images.

As we have presented, the unsupervised training of a parametric translation is a key technical contribution. 
It involves two carefully designed architectures with cycle consistency losses that make up for the lack of supervision in the form of any paired data. While our current implementation supports four interpertable maps, the design is scalable and can include additional maps, which in turn may unlock more control handles using advanced rendering setups.

Our \name, and neural rendering in general, questions \textit{ when do we really need 3D models?} In computer graphics, the production or the fabrication of a physical 3D model is of less important as the ultimate goal of the pipeline is to `merely' produce images.
Then, it is questionable whether the 3D representation is needed at all, or some abstract or latent representation could well instead be used. In our current work, we explored this avenue and reduced the requirement from detailed 3D models and/or explicit material assignment.  In the future, we would like to continue and search for new means to further reduce or totally avoid the use of explicit 3D, without losing any control over the quality or manipulation of the generated image. As the quality of neural image generation continue to rapidly improve, we believe that this work is an important step forward towards a fully neural workflow, without sacrificing users' ability to control the underlying scene attributes. 

\rev{
\section*{Acknowledgements}
We thank the anonymous reviewers for their insightful comments and feedback. This work is supported in part by grants from the National Key R\&D Program of China (2019YFF0302900, 2019YFF0302902), China Scholarship Council, Israel Science Foundation (grant no. 2366/16 and 2472/17), ERC Smart Geometry Grant, and gifts from Adobe.
}

\bibliographystyle{eg-alpha-doi}
\bibliography{ref}

\newcommand{\etalchar}[1]{$^{#1}$}
\begin{thebibliography}{\uppercase{GPAM{\etalchar{*}}14}}

\bibitem[ACB17]{wgan}
\textsc{Arjovsky M., Chintala S., Bottou L.}:
\newblock Wasserstein generative adversarial networks.
\newblock In \emph{International Conference on Machine Learning (ICML)} (2017),
  pp.~214--223.

\bibitem[ADMG17]{achlioptas2017learning}
\textsc{Achlioptas P., Diamanti O., Mitliagkas I., Guibas L.}:
\newblock Latent-space gans for {3D} point clouds.
\newblock In \emph{International Conference on Machine Learning (ICML),
  Workshop on Implicit Models} (2017).

\bibitem[ALS16]{dcgan}
\textsc{Alec R., Luke M., Soumith C.}:
\newblock Unsupervised representation learning with deep convolutional
  generative adversarial networks.
\newblock In \emph{International Conference on Learning Representations (ICLR)}
  (2016).

\bibitem[AUL19]{aliev2019neural}
\textsc{Aliev K.-A., Ulyanov D., Lempitsky V.}:
\newblock Neural point-based graphics.
\newblock \emph{arXiv preprint arXiv:1906.08240} (2019).

\bibitem[BDS19]{bigGAN}
\textsc{Brock A., Donahue J., Simonyan K.}:
\newblock Large scale {GAN} training for high fidelity natural image synthesis.
\newblock In \emph{International Conference on Learning Representations (ICLR)}
  (2019).

\bibitem[BSP{\etalchar{*}}19]{deepcg2real}
\textsc{Bi S., Sunkavalli K., Perazzi F., Shechtman E., Kim V.~G., Ramamoorthi
  R.}:
\newblock Deep cg2real: Synthetic-to-real translation via image
  disentanglement.
\newblock In \emph{International Conference on Computer Vision (ICCV)} (2019),
  pp.~2730--2739.

\bibitem[CFG{\etalchar{*}}15]{chang2015shapenet}
\textsc{Chang A.~X., Funkhouser T., Guibas L., Hanrahan P., Huang Q., Li Z.,
  Savarese S., Savva M., Song S., Su H., et~al.}:
\newblock Shapenet: An information-rich 3d model repository.
\newblock \emph{arXiv preprint arXiv:1512.03012} (2015).

\bibitem[Com]{blender}
\textsc{Community B.~O.}:
\newblock \emph{Blender - a 3D modelling and rendering package}.
\newblock Blender Foundation, Stichting Blender Foundation, Amsterdam.

\bibitem[CZ19]{im-net}
\textsc{Chen Z., Zhang H.}:
\newblock Learning implicit fields for generative shape modeling.
\newblock In \emph{Conference on Computer Vision and Pattern Recognition
  (CVPR)} (2019), pp.~5939--5948.

\bibitem[DDS{\etalchar{*}}09]{deng2009imagenet}
\textsc{Deng J., Dong W., Socher R., Li L.-J., Li K., Fei-Fei L.}:
\newblock Imagenet: A large-scale hierarchical image database.
\newblock In \emph{Conference on Computer Vision and Pattern Recognition
  (CVPR)} (2009), pp.~248--255.

\bibitem[EKJP20]{genesis}
\textsc{Engelcke M., Kosiorek A.~R., Jones O.~P., Posner I.}:
\newblock Genesis: Generative scene inference and sampling with object-centric
  latent representations.
\newblock \emph{International Conference on Learning Representations (ICLR)}
  (2020).

\bibitem[ERB{\etalchar{*}}18]{eslami2018neural_science}
\textsc{Eslami S.~A., Rezende D.~J., Besse F., Viola F., Morcos A.~S., Garnelo
  M., Ruderman A., Rusu A.~A., Danihelka I., Gregor K., et~al.}:
\newblock Neural scene representation and rendering.
\newblock \emph{Science 360}, 6394 (2018), 1204--1210.

\bibitem[GEB16]{gatys2016imagestyletransfer}
\textsc{Gatys L.~A., Ecker A.~S., Bethge M.}:
\newblock Image style transfer using convolutional neural networks.
\newblock In \emph{Conference on Computer Vision and Pattern Recognition
  (CVPR)} (2016), pp.~2414--2423.

\bibitem[GFK{\etalchar{*}}18]{atlasnet}
\textsc{Groueix T., Fisher M., Kim V.~G., Russell B., Aubry M.}:
\newblock {AtlasNet: A Papier-M\^ach\'e Approach to Learning 3D Surface
  Generation}.
\newblock In \emph{Conference on Computer Vision and Pattern Recognition
  (CVPR)} (2018).

\bibitem[GPAM{\etalchar{*}}14]{goodfellow2014generative}
\textsc{Goodfellow I., Pouget-Abadie J., Mirza M., Xu B., Warde-Farley D.,
  Ozair S., Courville A., Bengio Y.}:
\newblock Generative adversarial nets.
\newblock In \emph{Advances in Neural Information Processing Systems (NeurIPS)}
  (2014), pp.~2672--2680.

\bibitem[GYW{\etalchar{*}}19]{gao_siga19}
\textsc{Gao L., Yang J., Wu T., Fu H., Lai Y.-K., Zhang H.}:
\newblock {SDM-NET}: Deep generative network for structured deformable mesh.
\newblock \emph{ACM Transactions on Graphics (SIGGRAPH Asia) 38}, 6 (2019).

\bibitem[HLBK18]{huang2018multimodal}
\textsc{Huang X., Liu M.-Y., Belongie S., Kautz J.}:
\newblock Multimodal unsupervised image-to-image translation.
\newblock In \emph{European Conference on Computer Vision (ECCV)} (2018),
  pp.~172--189.

\bibitem[HRU{\etalchar{*}}17]{heusel2017gans}
\textsc{Heusel M., Ramsauer H., Unterthiner T., Nessler B., Hochreiter S.}:
\newblock Gans trained by a two time-scale update rule converge to a local nash
  equilibrium.
\newblock In \emph{Advances in Neural Information Processing Systems (NeurIPS)}
  (2017), pp.~6626--6637.

\bibitem[HZRS16]{resnet}
\textsc{He K., Zhang X., Ren S., Sun J.}:
\newblock Deep residual learning for image recognition.
\newblock In \emph{Conference on Computer Vision and Pattern Recognition
  (CVPR)} (2016), pp.~770--778.

\bibitem[KB15]{kingma2015adam}
\textsc{Kingma D.~P., Ba J.}:
\newblock Adam: A method for stochastic optimization.
\newblock In \emph{International Conference on Learning Representations (ICLR)}
  (2015).

\bibitem[KGS{\etalchar{*}}18]{KellyGuerreroEtAl_FrankenGAN_SigAsia2018}
\textsc{Kelly T., Guerrero P., Steed A., Wonka P., Mitra N.~J.}:
\newblock Frankengan: Guided detail synthesis for building mass models using
  style-synchonized gans.
\newblock \emph{ACM Transactions on Graphics (SIGGRAPH Asia) 37}, 6 (2018),
  1:1--1:14.

\bibitem[KHK18]{dtlcgan}
\textsc{Kaneko T., Hiramatsu K., Kashino K.}:
\newblock Generative adversarial image synthesis with decision tree latent
  controller.
\newblock In \emph{Conference on Computer Vision and Pattern Recognition
  (CVPR)} (2018).

\bibitem[KLA19]{stylegan}
\textsc{Karras T., Laine S., Aila T.}:
\newblock A style-based generator architecture for generative adversarial
  networks.
\newblock In \emph{Conference on Computer Vision and Pattern Recognition
  (CVPR)} (2019).

\bibitem[LKM{\etalchar{*}}18]{lucic2018gans}
\textsc{Lucic M., Kurach K., Michalski M., Gelly S., Bousquet O.}:
\newblock Are gans created equal? a large-scale study.
\newblock In \emph{Advances in Neural Information Processing Systems (NeurIPS)}
  (2018), pp.~700--709.

\bibitem[LXC{\etalchar{*}}17]{grass}
\textsc{Li J., Xu K., Chaudhuri S., Yumer E., Zhang H., Guibas L.}:
\newblock {GRASS}: Generative recursive autoencoders for shape structures.
\newblock \emph{ACM Transactions on Graphics (TOG) 36}, 4 (2017), Article 52.

\bibitem[MGK{\etalchar{*}}19]{neuralrerendering}
\textsc{Meshry M., Goldman D.~B., Khamis S., Hoppe H., Pandey R., Snavely N.,
  Martin-Brualla R.}:
\newblock Neural rerendering in the wild.
\newblock In \emph{Conference on Computer Vision and Pattern Recognition
  (CVPR)} (2019), pp.~6878--6887.

\bibitem[MGY{\etalchar{*}}19]{mo2019structurenet}
\textsc{Mo K., Guerrero P., Yi L., Su H., Wonka P., Mitra N., Guibas L.}:
\newblock Structurenet: Hierarchical graph networks for 3d shape generation.
\newblock \emph{ACM Transactions on Graphics (SIGGRAPH Asia) 38}, 6 (2019),
  Article 242.

\bibitem[MLX{\etalchar{*}}17]{lsgan}
\textsc{Mao X., Li Q., Xie H., Lau R.~Y., Wang Z., Paul~Smolley S.}:
\newblock Least squares generative adversarial networks.
\newblock In \emph{Proceedings of the IEEE International Conference on Computer
  Vision} (2017), pp.~2794--2802.

\bibitem[MO14]{conditionalGAN}
\textsc{Mirza M., Osindero S.}:
\newblock Conditional generative adversarial nets.
\newblock \emph{CoRR abs/1411.1784} (2014).

\bibitem[MPF{\etalchar{*}}18]{ma_siga18}
\textsc{Ma R., Patil A.~G., Fisher M., Li M., Pirk S., Hua B.-S., Yeung S.-K.,
  Tong X., Guibas L.~J., Zhang H.}:
\newblock Language-driven synthesis of 3d scenes using scene databases.
\newblock \emph{ACM Transactions on Graphics (SIGGRAPH Asia) 37}, 6 (2018).

\bibitem[NPLBY18]{rendernet}
\textsc{Nguyen-Phuoc T.~H., Li C., Balaban S., Yang Y.}:
\newblock Rendernet: A deep convolutional network for differentiable rendering
  from 3d shapes.
\newblock In \emph{Advances in Neural Information Processing Systems (NeurIPS)}
  (2018), pp.~7891--7901.

\bibitem[NPLT{\etalchar{*}}19]{hologan}
\textsc{Nguyen-Phuoc T., Li C., Theis L., Richardt C., Yang Y.-L.}:
\newblock Hologan: Unsupervised learning of 3d representations from natural
  images.
\newblock In \emph{International Conference on Computer Vision (ICCV)} (Nov
  2019).

\bibitem[NPRM{\etalchar{*}}20]{blockgan}
\textsc{Nguyen-Phuoc T., Richardt C., Mai L., Yang Y.-L., Mitra N.}:
\newblock Blockgan: Learning 3d object-aware scene representations from
  unlabelled images.
\newblock \emph{arXiv preprint arXiv:2002.08988} (2020).

\bibitem[OOS17]{acgan}
\textsc{Odena A., Olah C., Shlens J.}:
\newblock Conditional image synthesis with auxiliary classifier gans.
\newblock In \emph{International Conference on Machine Learning (ICML)} (2017),
  pp.~2642--2651.

\bibitem[OTW{\etalchar{*}}19]{tbn}
\textsc{Olszewski K., Tulyakov S., Woodford O., Li H., Luo L.}:
\newblock Transformable bottleneck networks.
\newblock \emph{International Conference on Computer Vision (ICCV)} (Nov 2019).

\bibitem[PFS{\etalchar{*}}19]{deepsdf}
\textsc{Park J.~J., Florence P., Straub J., Newcombe R., Lovegrove S.}:
\newblock Deepsdf: Learning continuous signed distance functions for shape
  representation.
\newblock In \emph{Conference on Computer Vision and Pattern Recognition
  (CVPR)} (June 2019).

\bibitem[PLWZ19]{spade}
\textsc{Park T., Liu M.-Y., Wang T.-C., Zhu J.-Y.}:
\newblock Semantic image synthesis with spatially-adaptive normalization.
\newblock In \emph{Conference on Computer Vision and Pattern Recognition
  (CVPR)} (2019), pp.~2337--2346.

\bibitem[RWL19]{ritchie2019fast}
\textsc{Ritchie D., Wang K., Lin Y.-a.}:
\newblock Fast and flexible indoor scene synthesis via deep convolutional
  generative models.
\newblock In \emph{Conference on Computer Vision and Pattern Recognition
  (CVPR)} (2019), pp.~6182--6190.

\bibitem[SLJ{\etalchar{*}}15]{szegedy2015going}
\textsc{Szegedy C., Liu W., Jia Y., Sermanet P., Reed S., Anguelov D., Erhan
  D., Vanhoucke V., Rabinovich A.}:
\newblock Going deeper with convolutions.
\newblock In \emph{Conference on Computer Vision and Pattern Recognition
  (CVPR)} (2015), pp.~1--9.

\bibitem[STH{\etalchar{*}}19]{deepvoxels}
\textsc{Sitzmann V., Thies J., Heide F., Nie{\ss}ner M., Wetzstein G.,
  Zollhofer M.}:
\newblock Deepvoxels: Learning persistent 3d feature embeddings.
\newblock In \emph{Conference on Computer Vision and Pattern Recognition
  (CVPR)} (2019), pp.~2437--2446.

\bibitem[SZA{\etalchar{*}}19]{neuralavatar}
\textsc{Shysheya A., Zakharov E., Aliev K.-A., Bashirov R., Burkov E., Iskakov
  K., Ivakhnenko A., Malkov Y., Pasechnik I., Ulyanov D., et~al.}:
\newblock Textured neural avatars.
\newblock In \emph{Conference on Computer Vision and Pattern Recognition
  (CVPR)} (2019), pp.~2387--2397.

\bibitem[SZW19]{neuralRepoNet_19}
\textsc{Sitzmann V., Zollh{\"o}fer M., Wetzstein G.}:
\newblock Scene representation networks: Continuous 3d-structure-aware neural
  scene representations.
\newblock In \emph{Advances in Neural Information Processing Systems (NeurIPS)}
  (2019), pp.~1119--1130.

\bibitem[TZN19]{thies2019deferred}
\textsc{Thies J., Zollh{\"o}fer M., Nie{\ss}ner M.}:
\newblock Deferred neural rendering: Image synthesis using neural textures.
\newblock \emph{ACM Transactions on Graphics (SIGGRAPH) 38}, 4 (2019), 1--12.

\bibitem[vSKG18]{obj_comp}
\textsc{van Steenkiste S., Kurach K., Gelly S.}:
\newblock A case for object compositionality in deep generative models of
  images.
\newblock \emph{CoRR abs/1810.10340} (2018).

\bibitem[WLW{\etalchar{*}}19]{wang2019planit}
\textsc{Wang K., Lin Y.-A., Weissmann B., Savva M., Chang A.~X., Ritchie D.}:
\newblock Planit: Planning and instantiating indoor scenes with relation graph
  and spatial prior networks.
\newblock \emph{ACM Transactions on Graphics (SIGGRAPH) 38}, 4 (2019), 1--15.

\bibitem[WLZ{\etalchar{*}}18]{wang2018high}
\textsc{Wang T.-C., Liu M.-Y., Zhu J.-Y., Tao A., Kautz J., Catanzaro B.}:
\newblock High-resolution image synthesis and semantic manipulation with
  conditional gans.
\newblock In \emph{Conference on Computer Vision and Pattern Recognition
  (CVPR)} (2018), pp.~8798--8807.

\bibitem[WSH{\etalchar{*}}19]{noc}
\textsc{Wang H., Sridhar S., Huang J., Valentin J., Song S., Guibas L.~J.}:
\newblock Normalized object coordinate space for category-level 6d object pose
  and size estimation.
\newblock In \emph{Conference on Computer Vision and Pattern Recognition
  (CVPR)} (June 2019).

\bibitem[WZX{\etalchar{*}}16]{3dgan}
\textsc{Wu J., Zhang C., Xue T., Freeman W.~T., Tenenbaum J.~B.}:
\newblock Learning a probabilistic latent space of object shapes via 3d
  generative-adversarial modeling.
\newblock In \emph{Advances in Neural Information Processing Systems (NeurIPS)}
  (2016), pp.~82--90.

\bibitem[YHH{\etalchar{*}}19]{pointflow}
\textsc{Yang G., Huang X., Hao Z., Liu M.-Y., Belongie S., Hariharan B.}:
\newblock Pointflow: 3d point cloud generation with continuous normalizing
  flows.
\newblock In \emph{International Conference on Computer Vision (ICCV)} (2019),
  pp.~4541--4550.

\bibitem[YJL{\etalchar{*}}18]{yang2018automatic}
\textsc{Yang Y., Jin S., Liu R., Bing~Kang S., Yu J.}:
\newblock Automatic 3d indoor scene modeling from single panorama.
\newblock In \emph{Conference on Computer Vision and Pattern Recognition
  (CVPR)} (2018), pp.~3926--3934.

\bibitem[YKBP17]{lrgan}
\textsc{Yang J., Kannan A., Batra D., Parikh D.}:
\newblock Lr-gan: Layered recursive generative adversarial networks for image
  generation.
\newblock \emph{International Conference on Learning Representations (ICLR)}
  (2017).

\bibitem[ZLB{\etalchar{*}}19]{terrainGAN19}
\textsc{Zhao Y., Liu H., Borovikov I., Beirami A., Sanjabi M., Zaman K.}:
\newblock Multi-theme generative adversarial terrain amplification.
\newblock \emph{ACM Transactions on Graphics (SIGGRAPH Asia) 38}, 6 (2019),
  200.

\bibitem[ZPIE17]{cyclegan}
\textsc{Zhu J.-Y., Park T., Isola P., Efros A.~A.}:
\newblock Unpaired image-to-image translation using cycle-consistent
  adversarial networks.
\newblock In \emph{International Conference on Computer Vision (ICCV)} (2017),
  pp.~2223--2232.

\bibitem[ZXC{\etalchar{*}}18]{zhu_siga18}
\textsc{Zhu C., Xu K., Chaudhuri S., Yi R., Zhang H.}:
\newblock {SCORES}: Shape composition with recursive substructure priors.
\newblock \emph{ACM Transactions on Graphics (SIGGRAPH Asia) 37}, 6 (2018).

\bibitem[ZYM{\etalchar{*}}18]{zhang2018deep}
\textsc{Zhang Z., Yang Z., Ma C., Luo L., Huth A., Vouga E., Huang Q.}:
\newblock Deep generative modeling for scene synthesis via hybrid
  representations.
\newblock \emph{arXiv preprint arXiv:1808.02084} (2018).

\bibitem[ZZZ{\etalchar{*}}18]{von}
\textsc{Zhu J.-Y., Zhang Z., Zhang C., Wu J., Torralba A., Tenenbaum J.,
  Freeman B.}:
\newblock Visual object networks: image generation with disentangled 3d
  representations.
\newblock In \emph{Advances in Neural Information Processing Systems (NeurIPS)}
  (2018), pp.~118--129.

\end{thebibliography}

\appendix

\section{Blinn-Phong rendering function}
\label{appendix:BP_function}
For rendering the final images, we assume that the reflectance of the content in the scene under camera view $v$ is characterized by a set of property maps: a surface normal map $\bm{N}$, a diffuse albedo map $\bm{I}_d$, a specular albedo map $\bm{I}_s$, and a monochrome specular roughness map $\bm{\alpha}$. We use a classical rendering model - \textit{Blinn-Phong Reflection Model} - as our rendering equation, which, for a given light $\bm{L}$, computes intensity as:
\begin{equation}
    \begin{gathered}
        \bm{I} = k_d (\bm{N} \cdot \bm{L}) \bm{I}_d + k_s (\bm{N} \cdot \bm{H})^{\bm{\alpha}} \bm{I}_s \\
        \bm{H} = \frac{ \bm{L} + \bm{V} }{ \norm{\bm{L} + \bm{V}} }
    \end{gathered}
    \label{eqn:BP}
\end{equation}
where$k_d$ and $k_s$ are diffuse reflection constant and specular reflection constant, respectively. $\bm{V}$ is the direction to the viewer, and hence is set to the view direction of $v$ for approximation. 

\section{Training losses for reflectance maps generation}
\label{appendix:losses}
We train the 2D networks for generating reflectance maps with a set of adversarial losses and cycle consistency losses. Each loss described in the following corresponds to a dashed arrow in the architecture figure in the main paper.
\paragraph{Adversarial losses}
For translating depth images to final composition images, we use the following adversarial loss for the detailed normal map generation:
\begin{equation} 
        \Lossnorm^{GAN} = \mathbb{E}_{\bm{N}^r}
        {\left[ \log \Dnorm(\bm{N}^r) \right]} 
        + 
        \mathbb{E}_{\bm{d}^r} {\left[ \log (1 - \Dnorm(\Gnorm(\bm{d}^r))  \right]}, 
    \label{eq:loss_normal_gan}
\end{equation}
where $\Dnorm$ learns to classify the real and generated normal maps. For the adversarial loss on the diffuse albedo maps generation:
\begin{equation}
    \begin{gathered}
        \Lossdiffa^{GAN} = \mathbb{E}_{\bm{I}_d^r}
        {\left[ \log \Ddiffa(\bm{I}_d^r) \right]} \\
        +
        \mathbb{E}_{(\bm{d}^r, \zdiffa)} {\left[ \log ( 1 - \Ddiffa( \Gdiffa(\dnoc^r, \Gnorm(\bm{{d}^r}), \zdiffa) ) )  \right]},
    \end{gathered}
    \label{eq:loss_diffuse_albedo_gan}
\end{equation}
where $\dnoc^r = \NOC(\bm{d}^r)$ and $\Ddiffa$ learns to classify the real and generated diffuse albedo maps. We also apply the adversarial loss on the diffuse images:
\begin{equation} 
    \begin{gathered}
        \Lossdiff^{GAN} = \mathbb{E}_{\bm{I}^r_{df}}
        {\left[ \log \Ddiff(\bm{I}^r_{df}) \right]} \\
        +
        \mathbb{E}_{\bm{d}^r} {\left[ \log ( 1 - \Ddiff( \Rdiff( \Gnorm(\bm{{d}^r}),
        \Gdiffa(\dnoc^r, \Gnorm(\bm{{d}^r}), \zdiffa),
        L) ) )  \right]}, 
    \end{gathered}
    \label{eq:loss_diffuse_gan}
\end{equation}
where $L$ is the light setting, $\bm{I}^r_{df}$ is the real diffuse image produced from real diffuse albedo and normal maps, and $\Ddiff$ learns to classify the real and generated diffuse images.

For the translation from diffuse images to depth images, we use the following adversarial loss:
\begin{equation} 
        \Lossdepth^{GAN} = \mathbb{E}_{\bm{d}^r}
        {\left[ \log \Ddepth(\bm{d}^r) \right]}
        + 
        \mathbb{E}_{\bm{I}^r_{df}} {\left[ \log (1 - \Ddepth( \Gdepth(\bm{I}^r_{df}) ))  \right]}, 
    \label{eq:loss_depth_gan}
\end{equation}
where $\Ddepth$ learns to classify the real and generated depth images. Furthermore, as we observed that the task of classifying the real depth images and generated ones is rather easier for $\Ddepth$, we also add the adversarial loss on the NOC image derived from the depth image to balance the network training:
\begin{equation} 
    \begin{gathered}
        \Lossnoc^{GAN} = \mathbb{E}_{\bm{d}^r}
            {\left[ \log \Dnoc(\NOC(\bm{d}^r)) \right]} \\
            + 
            \mathbb{E}_{\bm{I}^r_{df}} {\left[ \log (1 -\Dnoc( \NOC(\Gdepth(\bm{I}^r_{df})) )  \right]},
    \end{gathered}
    \label{eq:loss_noc_gan}
\end{equation}
where $\Dnoc$ learns to classify the real and generated NOC images.

\paragraph{Cycle-consistency losses}
We further add the following cycle consistency losses to enforce the bijective relationship between each two domains.

Cycle-consistency loss on the depth map:
\begin{equation}
    \begin{gathered}
        \Lossdepth^{cyc} = \\
        \mathbb{E}_{(\bm{d}^r, \zdiffa)} 
        {\left[ \norm{ \bm{d}^g - \bm{d}^r}_1 \right]},
    \end{gathered}
    \label{eq:loss_depth_cyc}
\end{equation}
where $\bm{d}^g = \Gdepth( \Rdiff(\Gnorm( \bm{d}^r ), \Gdiffa(\dnoc^r, \Gnorm( \bm{d}^r ), \zdiffa),  L) )$.

Cycle-consistency loss on the NOC map:
\begin{equation}
    \begin{gathered}
        \Lossnoc^{cyc} = \\
        \mathbb{E}_{(\bm{d}^r, \zdiffa)} 
        {\left[ \norm{ \dnoc^g - \dnoc^r}_1 \right]},
    \end{gathered}
    \label{eq:loss_noc_cyc}
\end{equation}
where $\dnoc^r = \NOC( \bm{d}^r )$ and $\dnoc^g = \NOC(\Gdepth( \Rdiff(\Gnorm( \bm{d}^r ), \Gdiffa(\dnoc^r, \Gnorm( \bm{d}^r ), \zdiffa),  L) ))$.

Cycle-consistency loss on the normal map:
\begin{equation}
    \begin{gathered}
        \Lossnorm^{cyc} =  \mathbb{E}_{(\bm{N}^r, \bm{I}_{d}^r)}
        {\left[ \norm{\Gnorm(\Gdepth( \Rdiff(\bm{N}^r, \bm{I}_{d}^r, L) )) - \bm{N}^r}_1 \right]};
    \end{gathered}
    \label{eq:loss_normal_cyc}
\end{equation}
And cycle-consistency loss on the diffuse albedo map:
\begin{equation}
    \begin{gathered}
        \Lossdiffa^{cyc} = \\
        \mathbb{E}_{(\bm{N}^r, \bm{I}_d^r, \bm{I}_{df}^r)} 
        {\left[ \norm{\Gdiffa(\Gdepth(\bm{I}^r_{df}), \bm{N}^r, \Ediffa(\bm{I}_d^r)) - \bm{I}_d^r}_1 \right]};
    \end{gathered}
    \label{eq:loss_diffuse_cyc}
\end{equation}
Cycle-consistency loss for the diffuse image:
\begin{equation}
    \begin{gathered}
        \Lossdiff^{cyc} = \\
        \mathbb{E}_{(\bm{N}^r, \bm{I}_d^r, \bm{I}_{df}^r)} 
        {\left[ \norm{ \bm{I}^g_{df}  - \bm{I}^r_{df} }_1 \right]},
    \end{gathered}
\end{equation}
where $\bm{I}^r_{df} = \Rdiff(\bm{N}^r, \bm{I}_d^r, L)$ and $\bm{I}^g_{df} = \Rdiff( \Gnorm(\Gdepth(\bm{I}^r_{df})), \Gdiffa(\Gdepth(\bm{I}^r_{df}), \bm{N}^r, \Ediffa(\bm{I}_d^r)),  L )$.

In addition, similar to the latent space reconstruction in other unconditional GANs and image-to-image translation works, we also introduce a latent space cycle-consistency loss to encourage $\Gdiffa$ to use the diffuse albedo code $\zdiffa$:
\begin{equation}
    \begin{gathered}
        \Losszdiffa^{cyc} =  \mathbb{E}_{(\bm{d}^r, \zdiffa )} 
        {\left[ \norm{ \Ediffa( \Gdiffa(\bm{d}^r_{noc}, \Gnorm(\bm{d}^r), \zdiffa) )
        - \zdiffa
        }_1 \right]}.
    \end{gathered}
    \label{eq:loss_diff_cyc}
\end{equation}

At last, to enable sampling at test time, we force $\Ediffa(\bm{I}_d^r)$ to be close to the standard Gaussian distribution, by adding a Kullback-Leibler (KL) loss on the $\zdiffa$ latent space: 
\begin{equation}
    \LossKL =  \mathbb{E}_{\bm{I}_d^r} \\
    {\left[ \mathcal{D}_{KL}(\Ediffa(\bm{I}_d^r) \Vert \GaussianDistribution ) \right]},
\end{equation}
where $\mathcal{D}_{KL}(p \Vert q) = - \int_z p(z) \log \frac{p(z)}{q(z)} dz$.

Finally, we write the final 2D modeling loss as:
\begin{equation}
    \begin{split}
        \LosstwoDmodeling & = \Lossnorm^{GAN} + \Lossdiffa^{GAN} + \Lossdiff^{GAN} + \Lossdepth^{GAN} + \Lossnoc^{GAN} \\
        & + \lambda_\text{n}^{cyc} \Lossnorm^{cyc} + \lambda_\text{da}^{cyc} \Lossdiffa^{cyc} + \lambda_\text{df}^{cyc} \Lossdiff^{cyc} \\
        & + \lambda_\text{depth}^{cyc} \Lossdepth^{cyc} + \lambda_\text{noc}^{cyc} \Lossnoc^{cyc}  + \lambda_{\zdiffa}^{cyc} \Losszdiffa^{cyc} + \lambda_\text{KL} \LossKL,
    \end{split}
    \label{eq:2D_modelling_loss}
\end{equation}
where $\lambda_\text{n}^{cyc}$, $\lambda_\text{da}^{cyc}$, $\lambda_\text{df}^{cyc}$, $\lambda_\text{depth}^{cyc}$, $\lambda_\text{noc}^{cyc}$,  $\lambda_{\zdiffa}^{cyc}$ and $\lambda_\text{KL}$ control the importance of each cycle consistency loss.

\section{Training details.}
The 3D generation network is trained as described in the original IM-NET paper. The $\zshape$ is sampled from the standard Gaussian distribution $\GaussianDistribution$, with the code dimension $\vert \zshape \vert = 200$. 
The generated implicit fields are converted to meshes by using $128 \times 128 \times 128$ grid samplings and Marching Cubes. 
The diffuse code $\zdiffa$ is also sampled from the standard Gaussian distribution $\GaussianDistribution$, with the code length $\vert \zdiffa \vert = 8$. We set the hyperparameters in Eq.~\ref{eq:2D_modelling_loss}  as,  $\lambda_\text{depth}^\text{cyc} = \lambda_\text{noc}^{cyc} = 10$, $\lambda_\text{n}^{cyc} = \lambda_\text{da}^{cyc} = \lambda_\text{df}^{cyc} = 25$, $\lambda_{\zdiffa}^{cyc} = 1$, $\lambda_\text{KL} = 0.001$. 
We use Adam optimizer~\cite{kingma2015adam} with a learning rate of 0.0001 for training all 2D networks.
We first train the reflectance maps generation networks for 300,000 samples, and then train the realistic specular generation networks for 200,000 samples, at last fine-tune the whole 2D setup by joint training.
The diffuse reflectance constant $k_d$ in Equation~\ref{eqn:BP} to 0.6 for cars and 0.8 for chairs.
At the inference time, the specular reflection constant $k_s$ in Equation~\ref{eqn:BP} is set to 0.4 for cars and 0.2 for chairs, if applicable.

\section{Details of datasets}
\label{appendix:datasets}
\paragraph{Real reflectance property map sets}
For training reflectance property map generators, we render each model in Blender to collect the real reflectance property maps. 
Each model is fit into a unit sphere placed at the origin.
The camera view is randomly sampled from the camera view distribution described next. 
For the dataset of real reflectance property maps, we random sample camera views and render the models in Blender, obtaining around 10k sets of reflectance property maps for car category and around 40k sets of reflectance property maps for chair category.

\paragraph{Camera view distribution}
We assume the camera is at a fixed distance of 2m to the origin and use a focal length of 50mm (35mm film equivalent). 
The camera location is restricted on a sphere, which can be parameterized as ($\rho = 2$, $\theta$, $\phi$), where $\theta$ is the counter-clockwise vertical angle from the object face-direction base and $\phi$ is the horizontal angle from the object face direction base. 
By default, we set the range of $\theta$ to be $\left[ \ang{0}, \ang{20} \right]$ and the range of $\phi$ to be $\left[ \ang{-90}, \ang{90} \right]$. 
In addition, we constrain the camera to look at the origin and disable camera in-plane rotation. 

\paragraph{Real images}
For training the realistic specular generator, we use the real-world images dataset from VON~\cite{von}, which contains 2605 car images and 1963 chair images. The images are randomly flipped during the training for data augmentation.

\section{Details of baseline methods and \name variants}
\label{appendix:baseline_details}
\paragraph{Baseline methods}
In the following, We describe the details of the baseline methods appeared in the comparison.
\begin{enumerate}[(i)]
    \item DCGAN~\cite{dcgan} proposed specific generator and discriminator architectures that significantly improve the training of generative adversarial networks. We use DCGAN with the standard cross-entropy loss.
    \item LSGAN~\cite{lsgan} adopted least square loss for stabilizing the GAN training. We use the same DCGAN generator and discriminator architectures for LSGAN.
    \item WGAN-GP~\cite{wgan} adopted Wasserstein metric and gradient penalty in training. We also use the same DCGAN generator and discriminator architectures for WGAN-GP. In addition, we replace the default BatchNorm by InstanceNorm in the discriminator, and train the discriminator 5 times per generator iteration.
    \item VON~\cite{von} also generates 3D rough shapes first but instead trains a network to add texture from a specific view to generate images. The VON results are obtained by the released models from the authors.
    \item SRNs~\cite{neuralRepoNet_19} formulates the image formation as a neural, 3D-aware rendering algorithm. SRNs assume having images with full camera parameters as training data, thus it can only be trained on composite images obtained by rendering the ShapeNet models using Blinn-Phong renderer. After trained, we make SRNs a generative model for image generation task by randomly \textit{pick} scene codes generated from the training and randomly sample camera viewpoints, similarly to the novel view synthesis application as described in the original paper. 
\end{enumerate}

\paragraph{\name variants}
In the following, We describe the details of NGP variants appeared in the paper.
\begin{enumerate}[(i)]
    \item \ours, as the default option, the final image is generated by blending the diffuse rendering of $\Rdiff$ under 4 base overhead lights (same setting as in training) with the realistic specular map. Note that only $\Rdiff$ is used to light the scene under the base lights, thus these base lights only result in diffuse reflection but no specular highlights in the final image.
    \item \oursbp, as a depleted option, where we use the full Blinn-Phong renderer $\Rbp$ under 4 base lights overhead (same setting as in training), along with randomly sampled lights, but \textit{without} blending with the realistic specular map.
    \item \oursplus, as a more versatile option that combines \ours and \oursbp for illumination control of additional lights. The output image of NGP is first formed, on top of which the diffuse reflection and specular reflection yielded by the additional lights via $\Rbp$ are added for producing the final image.
\end{enumerate}

\section{Evaluation on NGP variants}
\label{appendix:more_evaluations}

In table~\ref{table:ngp_variants}, we show the FID scores of the three NGP variants. We can see that, in general, all the three NGP variants consistently outperforms the other methods. \oursplus even yields slightly better results than \ours with additional illumination control. Interestingly, the \oursbp produces the best results on chairs even with a biased traditional rendering model (Blinn-Phong model). 
\begin{table}[h!]
    \centering
    \tiny
    \caption{FID comparison on real images data. Note that FIDs are computed against real images data.}
      
      \begin{tabular}{r|ccccccc}
        \toprule
            & DCGAN & LSGAN & WGAN-GP & VON & \oursbp  & \ours & \oursplus  \\
          \midrule
          car  & 130.5 & 171.4 & 123.4 & 83.3 &\textbf{ 67.2} & \textbf{58.3} & \textbf{54.8} \\
          \midrule
          chair & 225.0 & 225.3 & 184.9 & 51.8 & \textbf{47.9} & \textbf{51.2} & \textbf{50.3}  \\
         \bottomrule
        \end{tabular}%
\label{table:ngp_variants}
\end{table}

\section{Video results}
Please see the supplementary video for demonstration of camera/illumination control supported by \ours. Note that our generators, being view-specific, can lead to small changes across camera variations.

\rev{

\begin{figure}[h]
    \centering
	\includegraphics[width=0.48\textwidth]{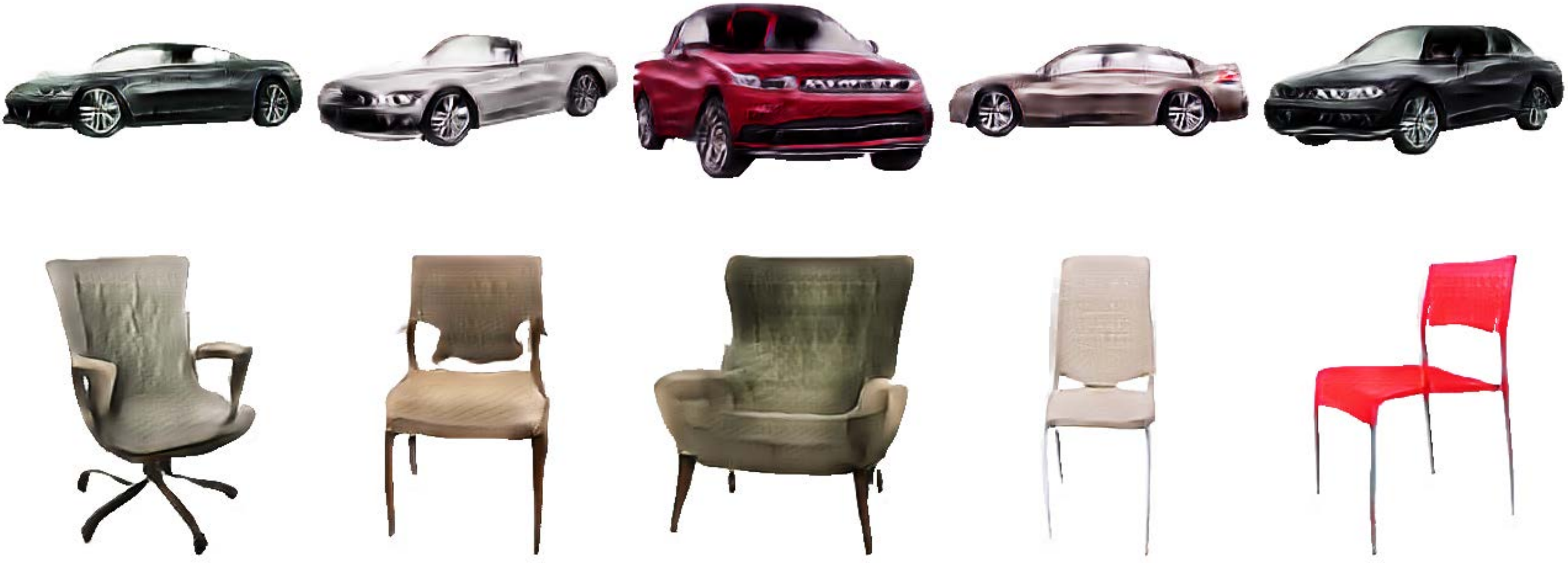}
	\caption{VON using our 3D shapes.}
    \label{fig:von_implicit_shapes}
\end{figure}

\section{VON using our 3D shapes}
We also show the results of training VON method, which directly operates in the image RGB space to texturize the 2D projections thus provides no control handle for the illumination, on our 3D shape data.
As it is non-trivial to adapt the original VON, particularly the differentiable 2D sketch generation module, to train with the implicit shapes used in our paper, we simply implement and train the core module --- the texturization networks, that texturize and translate the 2D sketch of the 3D shape into the final realistic image.
We can see that, in Figure~\ref{fig:von_implicit_shapes}, the visual results are similar to that in the original VON paper and no handle is available to control the illumination of the generated images.
}

\end{document}